\documentclass{article}

\usepackage[preprint]{neurips_2026}

\usepackage[utf8]{inputenc} 
\usepackage[T1]{fontenc}    
\usepackage{microtype}
\usepackage{graphicx}
\usepackage{subcaption}
\usepackage{booktabs} 
\usepackage{hyperref}
\usepackage{url}
\usepackage{amsfonts}
\usepackage{nicefrac}

\usepackage{amsmath}
\usepackage{amssymb}
\usepackage{mathtools}
\usepackage{amsthm}
\usepackage{multirow}
\usepackage{placeins} 
\usepackage{xcolor}
\usepackage{pifont} 
\usepackage{tikz}
\usepackage{pgfplots}
\pgfplotsset{compat=1.17}
\usetikzlibrary{shapes,arrows,positioning,calc,patterns,decorations.pathreplacing,fit,backgrounds,matrix}
\usepgfplotslibrary{groupplots,colorbrewer}
\definecolor{pythonblue}{RGB}{55,126,184}
\definecolor{esoRed}{RGB}{228,26,28}
\definecolor{esoOrange}{RGB}{255,127,0}
\definecolor{esoGreen}{RGB}{77,175,74}
\definecolor{esoPurple}{RGB}{152,78,163}
\definecolor{esoYellow}{RGB}{255,255,51}

\usepackage[capitalize,noabbrev]{cleveref}

\theoremstyle{plain}

\theoremstyle{definition}

\theoremstyle{remark}

\title{EsoLang-Bench: Evaluating Genuine Reasoning in Large Language Models via Esoteric Programming Languages}

\author{%
  Aman Sharma\thanks{Corresponding authors. Code: \url{https://github.com/Lossfunk/EsolangBench}. Dataset: \url{https://huggingface.co/datasets/arcAman07/Esolang-Bench}.} \\
  Lossfunk \\
  \texttt{aman.sharma@lossfunk.com} \\
  \And
  Paras Chopra\footnotemark[1] \\
  Lossfunk \\
  \texttt{paras@lossfunk.com} \\
}

\begin{document}

\begin{center}
\includegraphics[height=1.7cm]{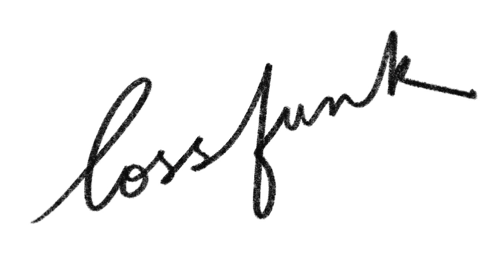}
\end{center}
\vspace{-0.5em}

\maketitle

\begin{abstract}
Large language models achieve near-ceiling performance on code generation benchmarks, yet most of the programming languages used by popular benchmarks such as SWE-bench and HumanEval (e.g.\ Python, JavaScript) are squarely in-distribution. They appear at scale in pre-training corpora and are heavily reinforced during post-training. To study LLM performance on unfamiliar programming languages, we introduce \textbf{EsoLang-Bench}, a benchmark using five esoteric programming languages (Brainfuck, Befunge-98, Whitespace, Unlambda, and Shakespeare). All five of our chosen esoteric languages are Turing-complete, so the same algorithmic problems that are solvable in Python or JavaScript are in principle solvable in each of them. Yet, they are unfamiliar to LLMs which makes them a good proxy for evaluating out-of-distribution performance. The unfamiliarity of esoteric languages comprises of: (i) the hard-by-design primitives comprising the language; (ii) substantially less representation in pre-training corpora (340$\times$ to over 60{,}000$\times$ fewer public GitHub repositories than Python); (iii) negligible deployment value, which makes targeted inclusion in post-training data economically irrational. We evaluate five frontier models across five prompting strategies and find a dramatic capability gap. The same 80 problems expressed in Python or JavaScript reach \textbf{100\%} accuracy on top frontier models, while the equivalent esoteric versions score only \textbf{0--11\%}. Few-shot learning and self-reflection also fail to close this gap. EsoLang-Bench therefore provides a contamination-resistant testbed for measuring how well frontier models generalise algorithmic problem-solving to programming languages outside their training distribution.
\end{abstract}

\section{Introduction}
\label{sec:intro}

Large language models have achieved impressive performance on code generation benchmarks, with state-of-the-art systems reaching 85-95\% accuracy on HumanEval and MBPP in mainstream programming languages \citep{chen2021humaneval, austin2021mbpp}. However, the languages those benchmarks target (Python, JavaScript, Java) are heavily represented in pre-training corpora and reinforced during post-training, and the benchmarks themselves increasingly suffer from contamination \citep{zhang2024gsm1k, sainz2023contamination} and surface-pattern shortcuts \citep{gupta2024benchmark}.

We do not have direct visibility into the training corpora of frontier closed models, so any \textit{out-of-distribution} (OOD) claim must be grounded in publicly observable proxies of pre-training scarcity. Esoteric programming languages offer a principled basis for such an evaluation. They have substantially fewer public repositories than mainstream languages (340$\times$ to over 60,000$\times$ fewer than Python depending on language, see Figure~\ref{fig:data-scarcity}), making large-scale presence in pre-training corpora highly unlikely. Within our five languages this scarcity itself spans two regimes that we will return to in the error analysis, a \emph{relatively-scarce} side (Brainfuck, Befunge-98) and a \emph{deeper-scarce} side (Whitespace, Unlambda, Shakespeare). Crucially, all five languages we evaluate are \emph{Turing-complete}, so the same algorithmic problems solvable in Python or JavaScript are also solvable in Brainfuck, Befunge-98, Whitespace, Unlambda, and Shakespeare. This decouples problem solvability (preserved by Turing-completeness) from familiarity to the model (eliminated by data scarcity).

\textbf{Limited practical utility is precisely the mechanism that makes the benchmark work.} These languages have negligible deployment value, so including them in post-training pipelines is economically unattractive at any meaningful scale, and combined with low pre-training presence this makes large-scale contamination unlikely at corpus scale rather than merely hoped against.

We make the following key contributions.
\begin{enumerate}
\setlength{\itemsep}{2pt}
\setlength{\parsep}{0pt}
    \item We introduce \textbf{EsoLang-Bench}, a dataset of 80 programming problems across four difficulty tiers and five esoteric programming languages with diverse computational paradigms and scarce pre-training data (340$\times$ to over 60{,}000$\times$ fewer GitHub repositories than Python), so the benchmark cannot be solved by surface-form memorisation of the target language.
    \item We conduct comprehensive experiments evaluating five state-of-the-art LLMs across five prompting strategies. Few-shot prompting adds only 0.8 percentage points over zero-shot on average, against an 89-point gap to mainstream baselines, and a pass@k baseline (Table~\ref{tab:passk}) corroborates that this small gain is a real capability ceiling rather than a single-sample artefact.
    \item We provide error analysis showing two qualitatively distinct failure regimes along the data-scarcity axis: a syntax-acquisition ceiling on the deepest-scarcity languages and a semantics-translation ceiling on the relatively-resourced ones, indicating that the bottleneck is foundational language familiarity rather than the iterative refinement mechanism layered on top (\S\ref{sec:error-analysis}).
\end{enumerate}

\begin{figure*}[t]
\centering
\includegraphics[width=\textwidth]{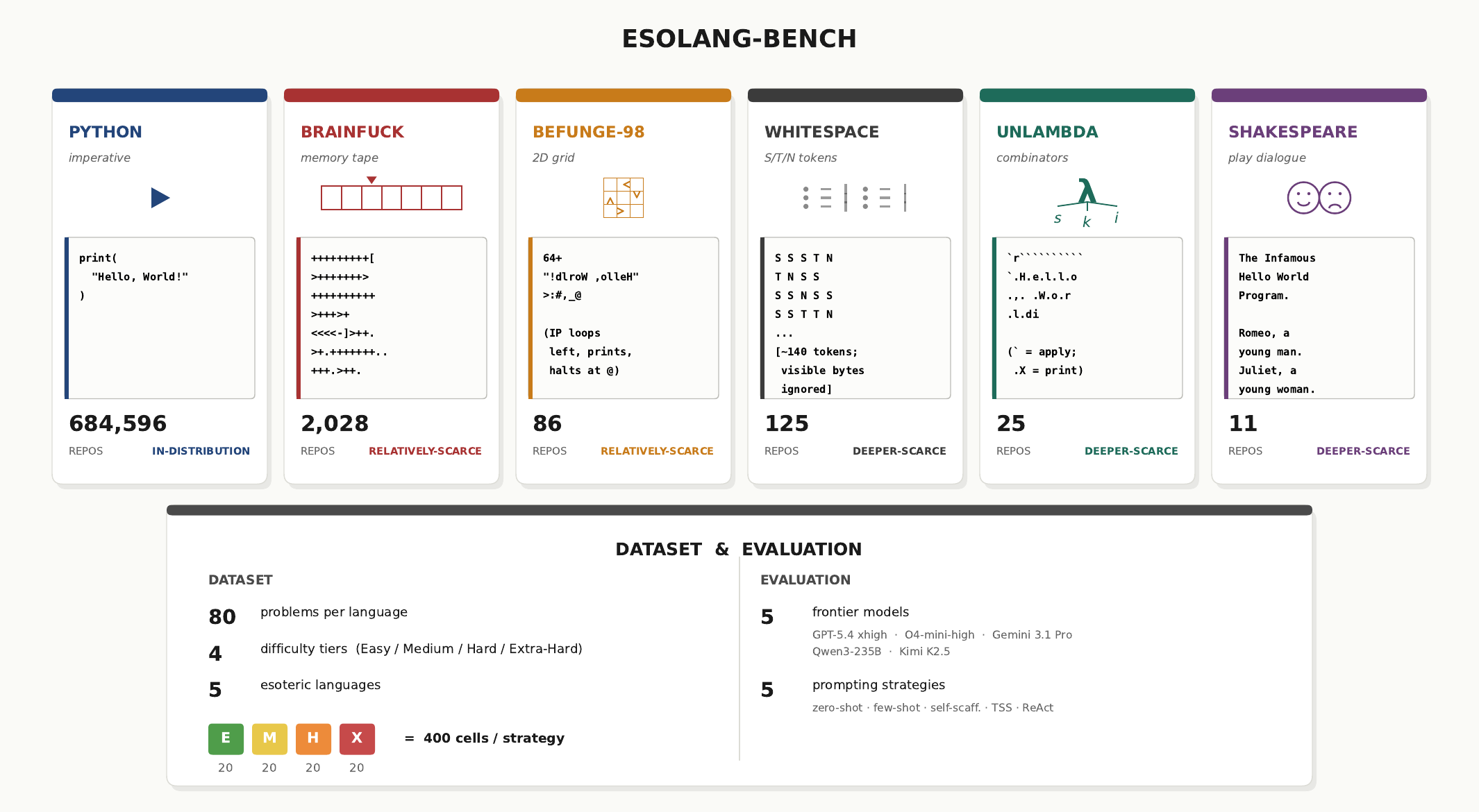}
\caption{\footnotesize\textbf{EsoLang-Bench at a glance.} \emph{Top:} the same Hello World written in Python and in each of the five esoteric languages targeted by the benchmark. \emph{Bottom:} the dataset (80 problems per language, 4 difficulty tiers, 5 esoteric languages) and the evaluation grid (5 frontier models $\times$ 5 prompting strategies).}
\label{fig:hello-world}
\end{figure*}

\section{Related work}

\subsection{Directly related work on esoteric-language ICL and contamination-resistant code benchmarks}
\label{sec:directly-related}

Most directly related is \citet{wang2025icl_esoteric}, which evaluates LLMs on Minipy, Pyth, Rhokell, and 0815. All four are Python-derived languages where ICL boosts Pyth accuracy from 16.7\% to 30.8\% by leveraging Python priors through near-identical surface syntax. Our five languages have entirely distinct computational models (memory tape, 2D grid, invisible whitespace, combinator calculus, theatrical natural-language), so demonstrations provide no analogous bridge and yield only 0.8 pp on average (\S\ref{sec:results}). The two papers are complementary. Their setting probes near-distribution reuse via syntactic recoding, while ours isolates far-OOD transfer when no such bridge exists. Among contamination-resistant code benchmarks, SWE-bench \citep{jimenez2024swebench} shows degradation on OOD repository tasks, and LiveCodeBench \citep{jain2025livecodebench} uses rolling release to preclude leakage by construction, paralleling our corpus-scarcity argument. \citet{liu2026convexbench} shows that frontier-model F1 drops from $\approx$1.0 at composition depth 2 to $\approx$0.2 at depth 100 on convex-function recognition, evidencing a sharp degradation under deep functional composition. This parallels our difficulty-tier design, where each tier raises algorithmic depth rather than statement length.

\subsection{Code generation benchmarks}

The evaluation of code generation capabilities has evolved through several generations of benchmarks. HumanEval \citep{chen2021humaneval} and MBPP \citep{austin2021mbpp} established the paradigm of function-level synthesis with test-case verification. SWE-bench \citep{jimenez2024swebench} extended this to repository-level tasks requiring understanding of complex codebases. MultiPL-E \citep{cassano2023multiple} broadened language coverage to 18 programming languages, while DS-1000 \citep{lai2023ds1000} focused specifically on data science tasks. AlphaCode \citep{li2022alphacode} demonstrated competitive performance on programming competitions.

Specialized code models have achieved impressive results on these benchmarks. StarCoder \citep{li2023starcoder}, Code Llama \citep{roziere2023codellama}, CodeGen \citep{nijkamp2023codegen}, InCoder \citep{fried2023incoder}, and CodeT5 \citep{wang2021codet5} represent the progression of code-specialized architectures. However, these benchmarks focus exclusively on mainstream languages with abundant pre-training and post-training coverage, leaving open the question of how well such performance transfers to programming languages outside the training distribution.

\subsection{Benchmark contamination and gaming}

The integrity of LLM evaluation has come under increasing scrutiny. \citet{zhang2024gsm1k} demonstrated systematic overperformance on GSM8k \citep{cobbe2021gsm8k} compared to the contamination-free GSM1k variant, with some models exhibiting accuracy gaps of up to 8\%. \citet{sainz2023contamination} found widespread contamination across NLP benchmarks. \citet{deng2024memorization} documented significant memorization affecting benchmark scores across model families, while \citet{zhou2023leakage} and \citet{xu2024leakage} quantified leakage patterns across benchmark families.

Most strikingly, \citet{gupta2024benchmark} revealed that simply changing the order of multiple-choice answers can decrease MMLU accuracy by up to 13\%, demonstrating that models exploit superficial patterns rather than understanding content. This phenomenon reflects Goodhart's Law \citep{goodhart1984monetary}: ``when a measure becomes a target, it ceases to be a good measure.'' \citet{strathern1997improving} extended this to academia, and we argue it applies directly to AI benchmarks.

\citet{jacovi2023contamination} proposed practical strategies for mitigating contamination, while \citet{oren2023contamination} developed methods for proving contamination in black-box models. \citet{bowman2021benchmarks} argued for fundamental reforms to NLU benchmarking, and \citet{raji2021benchmarks} critiqued the proliferation of narrow benchmarks. \citet{ribeiro2020behavioral} introduced behavioral testing methodologies that go beyond aggregate accuracy metrics.

\subsection{Out-of-distribution generalization}

Theoretical foundations for OOD generalization come from domain adaptation theory \citep{bendavid2010domain, ganin2016domain}, which establishes formal bounds on cross-domain transfer. Compositional generalization studies \citep{dziri2023compositional} reveal systematic failures in transformer architectures when faced with novel combinations of known primitives. \citet{anil2022limits} demonstrated failure of length generalization, while \citet{merrill2023expressive} analyzed the theoretical expressive power of transformers with chain-of-thought.

CrossFit \citep{ye2021crossfit} and Adversarial GLUE \citep{wang2021adversarial} studied robustness to distribution shift in NLP. \citet{chollet2019measure} introduced ARC-AGI, which measures intelligence through skill acquisition efficiency on out-of-distribution abstract reasoning tasks rather than in-distribution accuracy. \citet{liu2026convexbench} probes a similar OOD-composition axis on a symbolic mathematical surface. \citet{mitchell2021abstraction} surveys abstraction and reasoning capabilities. We extend this line of work by proposing esoteric programming languages as controlled probes for OOD generalisation in code generation, where mainstream-language baselines provide a directly measurable in-distribution reference point.

\subsection{Advanced prompting and reasoning}

In-context learning \citep{brown2020gpt3} demonstrated that large language models can adapt to new tasks through examples. Chain-of-thought prompting \citep{wei2022cot} and scratchpads \citep{nye2021scratchpad} improved reasoning by eliciting intermediate steps. Self-improvement methods including Self-Refine \citep{madaan2023selfrefine}, Reflexion \citep{shinn2023reflexion}, and self-debugging \citep{chen2023selfdebugging} enable iterative refinement through self-generated feedback.

ReAct \citep{yao2023react} synergizes reasoning with tool use, while program synthesis research \citep{gulwani2017program, ellis2021dreamcoder} provides theoretical foundations for inductive programming. Model-agnostic meta-learning \citep{finn2017maml} established frameworks for rapid adaptation. We systematically compare these approaches on OOD tasks (Section~\ref{sec:results}), revealing that advanced prompting strategies may amplify rather than bridge knowledge gaps when foundational understanding is absent.

\section{The EsoLang-Bench dataset}
\label{sec:dataset}

\subsection{Dataset design}

EsoLang-Bench consists of 80 programming problems organized into four difficulty levels of 20 problems each (Easy, Medium, Hard, Extra-Hard). Each problem ships with a natural-language description and 6 input-output test cases used by the evaluation harness only (the model's prompt contains only the description; test cases are released publicly with the dataset for transparency but never enter the prompt context). The candidate program is executed against every test case in the language's interpreter, the produced stdout is compared byte-for-byte to the expected stdout, and \textbf{the problem is declared solved if and only if all 6 test cases pass. There is no partial credit.} Problems are language-agnostic, and each can be implemented in any of the five target languages. The complete list of all 80 problems with full descriptions and test cases is provided in Appendix~\ref{app:dataset}.

Difficulty tiers are defined by the underlying algorithmic complexity of each problem, assessed from a language-agnostic pseudocode specification by expert judgement prior to any model evaluation. \textbf{Easy} problems are direct primitives such as integer addition (\texttt{a+b}), reading and echoing a line, or counting characters. These map to a single arithmetic, I/O, or loop operation. \textbf{Medium} problems require composing a few primitives, e.g.\ a Caesar cipher (per-character shift modulo 26), factorial via a counted loop, or run-length encoding. \textbf{Hard} problems require non-trivial algorithmic structure such as balanced-parentheses checking, prime counting up to $N$, or polynomial evaluation. \textbf{Extra-Hard} problems require classical algorithms with non-trivial state, e.g.\ the longest increasing subsequence, the Josephus problem, or counting inversions. Each tier deliberately raises algorithmic depth and not problem statement length, and Easy and Hard problems have comparable input sizes but very different reasoning requirements. A per-tier table of representative problems with pseudocode and asymptotic complexity is provided in Appendix~\ref{app:dataset}.

\textbf{The benchmark is hard for LLMs by construction.} Two distinct mechanisms compound. First, the target languages are under-represented in pre-training corpora and have negligible commercial deployment value, so they are unlikely to be heavily reinforced during post-training either. Second, and more importantly, the syntax of each language is deliberately exotic by design (hence the name esoteric), so even with full documentation in context, mapping an algorithm onto the language's instruction set requires non-trivial symbolic reasoning that does not benefit from the surface-syntax priors LLMs have absorbed for mainstream languages.

\subsection{Target languages and data scarcity}
\label{sec:data-scarcity}

We target five Turing-complete esoteric languages spanning distinct computational paradigms (full specifications in Appendix~\ref{app:languages}). The five are \textbf{Brainfuck} (8-command memory tape), \textbf{Befunge-98} (2D stack-based grid with directional instruction pointer), \textbf{Whitespace} (stack-based, only space, tab, newline carry meaning), \textbf{Unlambda} (combinatory logic via $s$, $k$, $i$), and \textbf{Shakespeare} (computation expressed as theatrical-play dialogue).

\begin{figure}[t]
\centering
\begin{tikzpicture}
\begin{axis}[
    xbar,
    width=0.98\columnwidth,
    height=5.4cm,
    xlabel={GitHub topic-tagged repositories ($\log_{10}$ scale)},
    xlabel style={font=\small},
    symbolic y coords={Shakespeare, Unlambda, Befunge-98, Whitespace, Brainfuck, JavaScript, Python},
    ytick=data,
    y tick label style={font=\small},
    x tick label style={font=\scriptsize},
    xmode=log,
    xmin=5, xmax=5000000,
    bar width=0.22cm,
    enlarge y limits=0.18,
    xtick={10, 100, 1000, 10000, 100000, 1000000},
    xticklabels={$10^1$, $10^2$, $10^3$, $10^4$, $10^5$, $10^6$},
    point meta=explicit symbolic,
    nodes near coords,
    every node near coord/.append style={anchor=west, font=\scriptsize, xshift=3pt},
]
\addplot[fill=pythonblue!70, draw=pythonblue!90] coordinates {
    (684596,Python)        [684{,}596]
    (648084,JavaScript)    [648{,}084]
    (2028,Brainfuck)       [2{,}028]
    (125,Whitespace)       [125]
    (86,Befunge-98)        [86]
    (25,Unlambda)          [25]
    (11,Shakespeare)       [11]
};
\end{axis}
\end{tikzpicture}
\caption{Per-language GitHub topic-tag counts on a $\log_{10}$ scale. Bar labels show the actual repository counts. Methodology and per-language details in Appendix~\ref{app:doc-completeness}.}
\label{fig:data-scarcity}
\end{figure}

Figure~\ref{fig:data-scarcity} shows the scarcity gap using GitHub topic-tag counts (verified May 2026). Python (684{,}596 repos) and JavaScript (648{,}084) tower over Brainfuck (2{,}028), Whitespace (125), Befunge-98 (86), Unlambda (25), and Shakespeare (11), a 340$\times$ to over 60{,}000$\times$ gap. Topic-tag counts under-estimate true repository counts uniformly, so the ratios are robust. Low pre-training presence combined with no commercial deployment value (which makes post-training inclusion economically unattractive) means benchmark gaming is highly unlikely at corpus scale. We address the natural objection that the gap reflects intrinsic language tractability rather than data scarcity in Appendix~\ref{app:tractability-objection}.

\section{Experimental setup}

\subsection{Models evaluated}

We evaluate five state-of-the-art LLMs representing the frontier of code generation capability. These are \textbf{GPT-5.4 xhigh}~\citep{openai2026gpt54}, \textbf{O4-mini-high}~\citep{openai2025o4} (OpenAI reasoning model), \textbf{Gemini 3.1 Pro}~\citep{gemini2026technical} (Google), \textbf{Qwen3-235B}~\citep{qwen3technical} (Alibaba), and \textbf{Kimi K2.5}~\citep{kimi25technical} (Moonshot). All models were accessed via API to ensure consistent evaluation conditions.

\subsection{Prompting strategies}

We evaluate five prompting strategies with increasing complexity, designed to systematically probe how different forms of scaffolding interact with out-of-distribution tasks.

\textbf{Zero-Shot.} The model receives only the language documentation and the problem description. The 6 input-output test cases are withheld from the prompt and used by the evaluation harness only. The system prompt establishes the model as an expert programmer and instructs it to output only valid code without explanations or markdown formatting. This baseline measures performance from documentation alone, without task-specific examples.

\textbf{Few-Shot.} Extends zero-shot by prepending 3 solved example programs in the target esoteric language, demonstrating correct syntax and I/O patterns. Examples are selected to cover basic constructs (loops, I/O, arithmetic) without revealing solutions to evaluation problems. This tests whether in-context learning can bridge knowledge gaps.

\textbf{Self-Scaffolding.} An iterative approach where the model generates code, receives interpreter feedback (actual vs. expected output, error messages, stack traces), and refines its solution for \textbf{up to 5 iterations}. Crucially, no separate critic model is involved, and the model must self-diagnose issues from raw interpreter output using a single LLM call per iteration. This isolates the benefit of execution feedback from explicit self-reflection. The exact format of the interpreter feedback the model receives at each iteration, with a worked example, is documented in Appendix~\ref{app:feedback-example}.

\textbf{Textual Self-Scaffolding.} A two-agent iterative process running for the \textbf{same up-to-5-iteration budget} as self-scaffolding and requiring two LLM API calls per iteration. The \textit{coder} generates code given the problem and any prior feedback, and a separate \textit{critic} agent analyzes the failing code and interpreter output to provide natural-language debugging guidance. The critic explicitly cannot write code and can only diagnose issues and suggest improvements. This tests whether externalized textual critique helps on OOD tasks.

\textbf{ReAct Pipeline.} A three-stage approach inspired by \citet{yao2023react}, also capped at \textbf{up to 5 iterations}. A \textit{planner} model generates a high-level algorithm in pseudocode, a \textit{code editor} translates the plan into the target esoteric language, and a \textit{critic} analyzes execution failures and feeds back to the planner. This tests whether decomposing reasoning and implementation helps when language-specific knowledge is missing.

\subsection{Evaluation protocol}
\label{sec:eval-protocol}

\textbf{Coverage.} We evaluate all five frontier models on the complete 80-problem dataset across all five esoteric languages, yielding 400 model-problem-language combinations per prompting strategy and 2{,}000 combinations per model when summed across the five strategies.

\textbf{Decoding parameters.} All primary evaluations use temperature $\tau = 0.2$ (close to deterministic to keep results reproducible) and a hard cap of \textbf{max\_tokens $= 32{,}000$ per response} so that even verbose esoteric programs (e.g.\ Whitespace and Shakespeare programs that are syntactically large) fit comfortably within a single generation. The pass@k baseline (Table~\ref{tab:passk}) is the only exception, and it uses $\tau = 0.8$ to obtain meaningful sample diversity across $n=8$ draws per problem.

\textbf{Inference stack.} All models are accessed through a single uniform API (OpenRouter) with web access disabled, and the only external context is the in-prompt material we provide (full language documentation in Appendix~\ref{app:doc-completeness}, three Easy-tier demonstrations for Few-Shot in Appendix~\ref{app:fewshot-examples}, and interpreter traces for the Self-Scaffolding family). The full sandboxed-execution model, error categorisation, per-strategy iteration budgets, retry policy, and statistical machinery are documented in Appendix~\ref{app:reproducibility}.

\section{Results}
\label{sec:results}

\begin{table*}[t]
\caption{Zero-shot and few-shot (3 ICL examples) accuracy (\%) across all models and languages. Each language contains 80 problems (20 per difficulty tier). Only Easy-tier problems were solved (all Medium/Hard/Extra-Hard = 0\%). Best results per language in \textbf{bold}. 0-S~=~Zero-Shot, 3-S~=~Three-Shot Few-Shot.}
\label{tab:main-results}
\vskip 0.1in
\begin{center}
\begin{tabular}{l@{\hskip 0.4cm}cc@{\hskip 0.4cm}cc@{\hskip 0.4cm}cc@{\hskip 0.4cm}cc@{\hskip 0.4cm}cc}
\toprule
 & \multicolumn{2}{c}{\textbf{Brainfuck}} & \multicolumn{2}{c}{\textbf{Befunge-98}} & \multicolumn{2}{c}{\textbf{Whitespace}} & \multicolumn{2}{c}{\textbf{Unlambda}} & \multicolumn{2}{c}{\textbf{Shakespeare}} \\
\cmidrule(lr){2-3} \cmidrule(lr){4-5} \cmidrule(lr){6-7} \cmidrule(lr){8-9} \cmidrule(lr){10-11}
\textbf{Model} & 0-S & 3-S & 0-S & 3-S & 0-S & 3-S & 0-S & 3-S & 0-S & 3-S \\
\midrule
GPT-5.4 xhigh        & 2.5\% & 2.5\% & 2.5\% & \textbf{8.8\%} & 0\% & 0\% & 0\% & 0\% & \textbf{2.5\%} & 1.2\% \\[0.08cm]
O4-mini             & 2.5\% & \textbf{3.8\%} & \textbf{6.2\%} & 7.5\% & 0\% & 0\% & 0\% & 0\% & 1.2\% & 1.2\% \\[0.08cm]
Gemini 3.1 Pro   & 2.5\% & 3.8\% & 5.0\% & 3.8\% & 0\% & 0\% & 0\% & 0\% & 1.2\% & 1.2\% \\[0.08cm]
Qwen-235B      & 2.5\% & 1.2\% & 0\% & 0\% & 0\% & 0\% & 0\% & \textbf{1.2\%} & 0\% & 0\% \\[0.08cm]
Kimi K2.5       & 0\% & 0\% & 2.5\% & 1.2\% & 0\% & 0\% & 0\% & 0\% & 1.2\% & 1.2\% \\
\bottomrule
\end{tabular}
\end{center}
\vskip -0.1in
\end{table*}

\begin{table*}[t]
\caption{Scaffolding strategy accuracy (\%) across all models and languages. S-S = Self-Scaffolding (direct interpreter feedback, 1 LLM call), TSS = Textual Self-Scaffolding (coder-critic pair, 2 LLM calls), Re = ReAct pipeline. Each language contains 80 problems. Best results in \textbf{bold}.}
\label{tab:scaffolding-results}
\vskip 0.1in
\begin{center}
\footnotesize
\begin{tabular}{l@{\hskip 0.18cm}ccc@{\hskip 0.18cm}ccc@{\hskip 0.18cm}ccc@{\hskip 0.18cm}ccc@{\hskip 0.18cm}ccc}
\toprule
 & \multicolumn{3}{c}{\textbf{Brainfuck}} & \multicolumn{3}{c}{\textbf{Befunge-98}} & \multicolumn{3}{c}{\textbf{Whitespace}} & \multicolumn{3}{c}{\textbf{Unlambda}} & \multicolumn{3}{c}{\textbf{Shakespeare}} \\
\cmidrule(lr){2-4} \cmidrule(lr){5-7} \cmidrule(lr){8-10} \cmidrule(lr){11-13} \cmidrule(lr){14-16}
\textbf{Model} & S-S & TSS & Re & S-S & TSS & Re & S-S & TSS & Re & S-S & TSS & Re & S-S & TSS & Re \\
\midrule
GPT-5.4 xhigh & \textbf{6.2} & 3.8 & 5.0 & \textbf{11.2} & 10.0 & 8.8 & 0 & 0 & 0 & \textbf{1.2} & 0 & 0 & \textbf{2.5} & 2.5 & 1.2 \\
O4-mini       & \textbf{5.0} & 2.5 & 3.8 & \textbf{10.0} & 6.2 & 7.5 & 0 & 0 & 0 & 0 & 0 & 0 & \textbf{1.2} & 1.2 & 0 \\
Gemini        & \textbf{5.0} & 3.8 & 3.8 & \textbf{7.5} & 6.2 & 7.5 & 0 & 0 & 0 & 0 & 0 & 0 & \textbf{1.2} & 0 & 0 \\
Qwen-235B     & \textbf{2.5} & 1.2 & 2.5 & 0 & 0 & 0 & 0 & 0 & 0 & \textbf{1.2} & 0 & 0 & \textbf{1.2} & 0 & 0 \\
Kimi K2.5     & 0 & 0 & 0 & 0 & 0 & 0 & 0 & 0 & 0 & 0 & 0 & 0 & \textbf{1.2} & 0 & 0 \\
\bottomrule
\end{tabular}
\end{center}
\vskip -0.05in
\textit{All values are accuracy in \%.}
\vskip -0.1in
\end{table*}

Tables~\ref{tab:main-results} and~\ref{tab:scaffolding-results} present our main experimental results. Accuracy is computed as the percentage of problems solved out of 80 total problems per language (20 Easy + 20 Medium + 20 Hard + 20 Extra-Hard).

\subsection{Mainstream-language baselines (Python and JavaScript)}
\label{sec:mainstream-baselines}

To isolate the OOD source of difficulty from intrinsic problem hardness, we ran the same 80 problems through the same pipeline using Python and JavaScript, evaluated under the same harness on the \emph{same five frontier models} used throughout the rest of this paper (GPT-5.4 xhigh, O4-mini-high, Gemini 3.1 Pro, Qwen3-235B, Kimi K2.5). Top frontier models reach 100\% on every difficulty tier in either language and zero-shot equals few-shot for every cell (full per-tier breakdown in Appendix~\ref{app:mainstream-baselines}). Figure~\ref{fig:mainstream-vs-eso} shows the resulting 89-point collapse from 100\% on Python or JavaScript to 11.2\% on the most tractable esoteric language (Befunge-98 under self-scaffolding), confirming that the esoteric-language results reflect leaving the pre-training distribution rather than broken tasks or a broken evaluation environment.

\begin{figure}[h]
\centering
\begin{tikzpicture}
\begin{axis}[
    xbar,
    width=0.88\columnwidth,
    height=5.0cm,
    xlabel={Accuracy (\%)},
    xlabel style={font=\small},
    symbolic y coords={
        Whitespace,Unlambda,Shakespeare,Brainfuck,Befunge-98,JavaScript,Python},
    ytick=data,
    y tick label style={font=\small},
    x tick label style={font=\small},
    xmin=0, xmax=110,
    bar width=0.22cm,
    enlarge y limits=0.16,
    nodes near coords,
    nodes near coords style={font=\scriptsize},
    nodes near coords align={horizontal},
]
\addplot[fill=pythonblue!75, draw=pythonblue!95] coordinates {
    (100,Python)
    (100,JavaScript)
    (11.2,Befunge-98)
    (6.2,Brainfuck)
    (2.5,Shakespeare)
    (1.2,Unlambda)
    (0,Whitespace)
};
\end{axis}
\end{tikzpicture}
\caption{Best-of-models accuracy on the same 80 EsoLang-Bench problems expressed in mainstream vs.\ esoteric languages. Mainstream baselines (Python, JavaScript) reach 100\% (top frontier models solve every problem), while esoteric peaks are language-specific best-of across all five prompting strategies.}
\label{fig:mainstream-vs-eso}
\end{figure}

\subsection{Static prompting performance}

In Table~\ref{tab:main-results}, Befunge-98 achieves the highest accuracy, followed by Brainfuck and Shakespeare. All models achieve 0\% on Whitespace and near-zero on Unlambda, marking a sharp capability boundary.

Within the Easy tier, top models solve a substantial fraction of problems. GPT-5.4 xhigh self-scaffolding solves 9/20 Easy Befunge-98 problems (45\%) and 5/20 Easy Brainfuck problems (25\%). This tier-level view reveals a sharper empirical story than aggregate accuracy suggests, since models are not uniformly failing but exhibit a hard performance cliff precisely at the boundary between single-loop pattern mapping (Easy) and multi-step algorithmic reasoning at higher tiers.

\textbf{Few-shot prompting adds at most 0.8 pp over zero-shot.} The mean cross-language lift from prepending three solved demonstrations is $+0.8$ pp, against an 89-point gap between mainstream and esoteric performance, and a paired Wilcoxon signed-rank test on per-problem solve indicators returns $p = 0.505$ (n.s.) for few-shot versus zero-shot. We additionally rank models by pass@k as a secondary baseline (Table~\ref{tab:passk}, $n=8$ draws, $T=0.8$, zero-shot). Cells at 0\% under single-sample evaluation remain at 0\% across all 8 draws on every model, and the strongest cell rises only modestly (e.g.\ GPT-5.4 xhigh on Befunge-98 from 5.0\% to 8.8\%, with a 95\% Clopper-Pearson upper bound of $\sim$0.32 on the true pass rate, far below the 89-point gap). Few-shot prompting therefore cannot bridge the OOD gap on this benchmark, consistent with the prediction of \citet{min2022icl} that in-context learning depends heavily on pre-training-distribution coverage. The few-shot examples are described in Appendix~\ref{app:fewshot-examples}.

\textbf{Forward-compatibility.} The Easy tier already differentiates frontier systems, with five-model variance of 0--11.2\% on Befunge-98 self-scaffolding (the most tractable esoteric cell), while every Medium / Hard / Extra-Hard cell and Whitespace and Unlambda remain unsolved by every model. The benchmark stays diagnostic across model generations as harder tiers and deeper-scarcity languages unlock progressively.

\begin{table}[t]
\caption{Zero-shot pass@$k$ (\%, of 80; $n=8$ draws, $T=0.8$) for $k\in\{1,8\}$. BF=Brainfuck, BE=Befunge-98, WS=Whitespace, UN=Unlambda, SH=Shakespeare. WS, UN, and SH stay at $\sim$0\% for every model and every $k$, and all Medium/Hard/Extra-Hard remain unsolved at any $k$.}
\label{tab:passk}
\vskip 0.05in
\begin{center}
\small
\begin{tabular}{l@{\hskip 0.2cm}cc@{\hskip 0.3cm}cc@{\hskip 0.3cm}cc@{\hskip 0.3cm}cc@{\hskip 0.3cm}cc@{\hskip 0.3cm}cc}
\toprule
 & \multicolumn{2}{c}{BF} & \multicolumn{2}{c}{BE} & \multicolumn{2}{c}{WS} & \multicolumn{2}{c}{UN} & \multicolumn{2}{c}{SH} & \multicolumn{2}{c}{Avg} \\
\cmidrule(lr){2-3}\cmidrule(lr){4-5}\cmidrule(lr){6-7}\cmidrule(lr){8-9}\cmidrule(lr){10-11}\cmidrule(lr){12-13}
\textbf{Model} & 1 & 8 & 1 & 8 & 1 & 8 & 1 & 8 & 1 & 8 & 1 & 8 \\
\midrule
GPT-5.4 xhigh            & 2.5 & 3.8 & 5.0 & 8.8 & 0.0 & 0.0 & 0.0 & 0.0 & 0.0 & 0.0 & 1.5 & 2.5 \\
O4-mini-high       & 2.5 & 2.5 & 5.0 & 7.5 & 0.0 & 0.0 & 0.0 & 1.2 & 0.0 & 0.0 & 1.5 & 2.2 \\
Gemini 3.1 Pro     & 2.5 & 3.8 & 5.0 & 5.0 & 0.0 & 0.0 & 0.0 & 0.0 & 0.0 & 0.0 & 1.5 & 1.8 \\
Qwen3-235B         & 1.2 & 2.5 & 2.5 & 3.8 & 0.0 & 0.0 & 0.0 & 0.0 & 0.0 & 0.0 & 0.8 & 1.2 \\
Kimi K2.5           & 1.2 & 2.5 & 3.8 & 5.0 & 0.0 & 0.0 & 0.0 & 0.0 & 0.0 & 0.0 & 1.0 & 1.5 \\
\bottomrule
\end{tabular}
\end{center}
\end{table}

\subsection{Scaffolding strategy performance}

Table~\ref{tab:scaffolding-results} presents results for three iterative scaffolding approaches. Self-scaffolding yields the best overall result with GPT-5.4 xhigh reaching 11.2\% on Befunge-98, an order of magnitude above zero-shot for that cell. Textual self-scaffolding (TSS) tracks self-scaffolding closely with cross-language Spearman $\rho = 0.94$ (95\% bootstrap CI [0.87, 0.98], 1{,}000 resamples) on per-cell solve counts, despite using twice the compute per iteration, and a paired Wilcoxon test returns $p = 0.803$ (n.s.) for TSS versus self-scaffolding, so direct interpreter feedback already captures the benefit and the textual critic adds nothing. ReAct also tracks self-scaffolding closely. \textbf{Self-scaffolding remains the strongest non-trivial strategy at half the compute of TSS and ReAct}, and the headline 11.2\% Befunge-98 result is attained under it. This pattern suggests the benefit derives from direct interpreter feedback rather than the textual critique mechanism, since for OOD tasks concrete execution traces (actual vs.\ expected output, exact error message, exact crash position) provide a sharper learning signal than another model's interpretation of the same failure, and on languages where the critic itself lacks domain knowledge a textual intermediary introduces additional noise rather than additional signal.

\section{Error analysis}
\label{sec:error-analysis}

\begin{figure}[tb]
\centering
\begin{tikzpicture}
\begin{axis}[
    ybar stacked,
    bar width=0.55cm,
    width=\columnwidth,
    height=4.3cm,
    ylabel={Error distribution (\%)},
    ylabel style={font=\small},
    symbolic x coords={BF, Bef, WS, Unl, Shk},
    xtick=data,
    x tick label style={font=\small},
    ymin=0, ymax=100,
    ytick={0,25,50,75,100},
    y tick label style={font=\small},
    legend style={at={(0.5,-0.20)}, anchor=north, legend columns=3, font=\small},
    enlarge x limits=0.13,
]
\addplot[fill=esoRed!80] coordinates {(BF,15) (Bef,20) (WS,100) (Unl,90) (Shk,70)};
\addplot[fill=esoYellow!80] coordinates {(BF,25) (Bef,45) (WS,0) (Unl,5) (Shk,5)};
\addplot[fill=esoGreen!70] coordinates {(BF,60) (Bef,35) (WS,0) (Unl,5) (Shk,25)};
\legend{Compile, Runtime, Logic}
\end{axis}
\end{tikzpicture}
\caption{Error distribution by language (GPT-5.4 xhigh, zero-shot). BF=Brainfuck, Bef=Befunge-98, WS=Whitespace, Unl=Unlambda, Shk=Shakespeare. Whitespace and Unlambda show near-total compile failure, and Brainfuck and Befunge-98 are dominated by logic errors despite producing parsable code. Per-model error breakdowns are in Appendix~\ref{app:error-analysis}.}
\label{fig:error-dist}
\end{figure}

Figure~\ref{fig:error-dist} decomposes GPT-5.4 xhigh zero-shot failures into three categories. \textbf{Compile errors} are syntactically invalid output. \textbf{Runtime errors} parse but crash mid-execution. \textbf{Logic errors} execute cleanly but stdout mismatches the expected output. Two qualitatively different failure regimes appear along the data-scarcity axis.

\textbf{Logic errors dominate on the relatively-resourced side (Brainfuck, Befunge-98).} These are the two highest-scarcity-but-still-non-trivial languages by GitHub topic-tag count (Brainfuck 2{,}028 repos, Befunge-98 86; Figure~\ref{fig:data-scarcity}). Compile errors are 15--20\% and logic errors 35--60\%, so the model produces parsable code but implements the wrong algorithm. Inspection shows it typically recognises the algorithmic shape but mis-translates into the restricted instruction set, e.g.\ off-by-one Brainfuck pointer arithmetic or wrong Befunge-98 instruction-pointer direction at a branching cell.

\textbf{Compile failure dominates on the deeper-scarcity side (Whitespace, Unlambda, Shakespeare).} These are the three lowest-corpus-presence languages in our set (Whitespace 125, Unlambda 25, Shakespeare 11 repos; Figure~\ref{fig:data-scarcity}). 70--100\% of attempts fail before execution starts and logic errors are essentially absent because the model rarely produces parsable code in the first place. Whitespace generations frequently include visible characters that the interpreter rejects on parse (recall that only space, tab, and newline are semantically valid). Shakespeare generations violate the stage-direction grammar at the persona-declaration step, e.g.\ failing to introduce characters before they appear in dialogue. Unlambda generations misuse the combinator application syntax. In each case the model has not internalised the language's syntactic constraints from the in-prompt documentation alone.

The transition between these two regimes is a categorical shift in which kind of failure dominates rather than a smooth accuracy degradation, which suggests that the OOD ceiling has two empirically distinguishable signatures rather than acting as a single uniform bottleneck. On the deepest-scarcity languages, the model fails at the syntax-acquisition step. On the relatively-resourced languages, the model has acquired enough surface form to produce parsable code, and the failure shifts to the semantic step of translating the algorithm into the language's restricted instruction set. The two failure modes are different in kind, so reading the headline accuracy as a single number hides this structure. The Whitespace 0\% result is examined in Appendix~\ref{app:whitespace-zero}, which rules out tokeniser and prompt-engineering artefacts and reads the result as consistent with corpus scarcity and the absence of post-training incentive to cover the language.

\section{Limitations}
\label{sec:limitations}

We acknowledge two limitations. \textbf{(i)} Error classification is coarse (compile, runtime, logic), and finer-grained failure-mode analysis (e.g.\ off-by-one pointer arithmetic, wrong instruction-pointer direction, mis-encoded immediate operands) would yield deeper diagnostics. \textbf{(ii)} The benchmark currently contains 80 problems per language. This was a planning choice rather than a budget constraint, since accuracy does not saturate even on the easier tiers and additional problems would not yet change the qualitative conclusion. We mitigate the moderate per-cell resolution via paired Wilcoxon tests, $n=1{,}000$ bootstrap, and the pass@k ($n=8$) bounding study, and will continue to expand the benchmark over time. We also note that most Medium, Hard, and Extra-Hard cells remain at 0\% across all five models and all five strategies, so present discriminative power is concentrated on the Easy tier of Brainfuck, Befunge-98, and Shakespeare, and we expect this floor to lift as frontier capability grows. Maintenance plans and the contribution process for new languages are in Appendix~\ref{app:datasheet}.

\section{Conclusion}
\label{sec:conclusion}

We introduced \textbf{EsoLang-Bench}, a contamination-resistant code-generation benchmark on five Turing-complete esoteric languages (Brainfuck, Befunge-98, Whitespace, Unlambda, Shakespeare) that are 340$\times$ to over 60{,}000$\times$ less represented on GitHub than Python and that span a \emph{relatively-scarce} regime (Brainfuck, Befunge-98) and a \emph{deeper-scarce} regime (Whitespace, Unlambda, Shakespeare) under which the dominant failure mode flips categorically from logic to compile (\S\ref{sec:error-analysis}). Mainstream-language baselines reach \textbf{100\%} on the same 80 problems while \textbf{peak esoteric accuracy is 11.2\%}, an 89-point collapse on identical algorithmic content, and few-shot prompting adds only \textbf{0.8 percentage points} over zero-shot. As frontier models saturate mainstream code benchmarks, reliable capability measurement needs evaluation surfaces orthogonal to commercial incentives, and we intend EsoLang-Bench as one realisation of that principle and a template for future OOD benchmarks.

\bibliography{esolang_bench}
\bibliographystyle{plainnat}

\newpage
\appendix

\section{Schematic overview of the benchmark and evaluation pipeline}
\label{app:hero-schematic}

Figure~\ref{fig:hero} provides a schematic overview of the benchmark and the evaluation pipeline. It is reproduced here as supplementary context to Figure~\ref{fig:hello-world} of the main paper.

\begin{figure*}[h]
\centering
\begin{tikzpicture}[scale=0.92, every node/.style={transform shape}]

\fill[blue!15, rounded corners=8pt] (-8.2, -3.6) rectangle (1.8, 3.7);
\node[font=\bfseries\large, text=blue!70!black] at (-3.2, 3.2) {EsoLang-Bench};

\node[circle, fill=blue!50, text=white, font=\bfseries\small, minimum size=0.55cm] at (-7.8, 2.4) {1};

\node[font=\bfseries\small, text=blue!70!black] at (-3.2, 2.4) {Target Languages};

\begin{scope}[shift={(-3.2, 1.1)}]
\node[draw=esoRed!60, rounded corners, fill=esoRed!15, minimum width=2.2cm, minimum height=0.6cm, align=center, font=\tiny] at (-2.4, 0.4) {\textbf{Brainfuck}\\Memory Tape};
\node[draw=esoOrange!60, rounded corners, fill=esoOrange!15, minimum width=2.2cm, minimum height=0.6cm, align=center, font=\tiny] at (0, 0.4) {\textbf{Befunge-98}\\2D Grid};
\node[draw=esoYellow!80, rounded corners, fill=esoYellow!25, minimum width=2.2cm, minimum height=0.6cm, align=center, font=\tiny] at (2.4, 0.4) {\textbf{Whitespace}\\Stack-Based};
\node[draw=esoGreen!60, rounded corners, fill=esoGreen!15, minimum width=2.2cm, minimum height=0.6cm, align=center, font=\tiny] at (-1.2, -0.35) {\textbf{Unlambda}\\Combinators};
\node[draw=esoPurple!60, rounded corners, fill=esoPurple!15, minimum width=2.2cm, minimum height=0.6cm, align=center, font=\tiny] at (1.2, -0.35) {\textbf{Shakespeare}\\Natural-Lang};
\end{scope}

\node[circle, fill=blue!50, text=white, font=\bfseries\small, minimum size=0.55cm] at (-7.8, -0.5) {2};

\node[font=\bfseries\small, text=blue!70!black] at (-3.2, -0.5) {Dataset Structure};

\begin{scope}[shift={(-3.2, -1.9)}]
\fill[green!10, rounded corners=4pt] (-2.9, -1.4) rectangle (2.9, 0.7);
\node[font=\footnotesize\bfseries] at (0, 0.45) {80 Problems $\times$ 5 Languages};
\draw[gray!50] (-2.9, 0.15) -- (2.9, 0.15);
\node[font=\scriptsize, fill=green!25, rounded corners, inner sep=2pt] at (-1.5, -0.2) {Easy: 20};
\node[font=\scriptsize, fill=yellow!35, rounded corners, inner sep=2pt] at (1.5, -0.2) {Medium: 20};
\node[font=\scriptsize, fill=orange!35, rounded corners, inner sep=2pt] at (-1.5, -0.7) {Hard: 20};
\node[font=\scriptsize, fill=red!25, rounded corners, inner sep=2pt] at (1.5, -0.7) {X-Hard: 20};
\node[font=\scriptsize\bfseries] at (0, -1.15) {400 Total Evaluations};
\end{scope}

\fill[orange!15, rounded corners=8pt] (2.2, -3.6) rectangle (8.2, 3.7);
\node[font=\bfseries\large, text=orange!70!black] at (5.2, 3.2) {Evaluation Pipeline};

\node[draw=orange!50, rounded corners, fill=white, minimum width=4.3cm, minimum height=0.85cm, align=center, font=\small] (models) at (5.2, 2.2) {\textbf{Frontier Models}\\{\tiny GPT-5.4 xhigh, O4-mini, Gemini, Qwen, Kimi}};
\draw[->, thick, orange!70] (5.2, 1.7) -- (5.2, 1.3);

\node[draw=orange!50, rounded corners, fill=white, minimum width=4.3cm, minimum height=0.85cm, align=center, font=\small] (prompts) at (5.2, 0.85) {\textbf{Prompting Strategies}\\{\tiny Zero-Shot, Few-Shot, Self-Scaffolding, ReAct}};
\draw[->, thick, orange!70] (5.2, 0.35) -- (5.2, -0.05);

\node[draw=orange!50, rounded corners, fill=white, minimum width=4.3cm, minimum height=0.7cm, align=center, font=\small] (codegen) at (5.2, -0.55) {\textbf{Code Generation}};
\draw[->, thick, orange!70] (5.2, -0.95) -- (5.2, -1.35);

\node[draw=orange!50, rounded corners, fill=white, minimum width=4.3cm, minimum height=0.7cm, align=center, font=\small] (exec) at (5.2, -1.85) {\textbf{Interpreter Execution}};
\draw[->, thick, orange!70] (5.2, -2.25) -- (5.2, -2.65);

\node[draw=orange!60, rounded corners, fill=orange!25, minimum width=4.3cm, minimum height=0.7cm, align=center, font=\small] (results) at (5.2, -3.1) {\textbf{Peak} 11.2\% vs.\ \textbf{Python} 100\%};

\end{tikzpicture}
\caption{\textbf{EsoLang-Bench overview (schematic).} \textit{Left.} The benchmark comprises five esoteric programming languages spanning diverse computational paradigms, with 80 problems across four difficulty tiers (400 total evaluations). \textit{Right.} The evaluation pipeline tests five frontier models across five prompting strategies (zero-shot, few-shot, self-scaffolding, textual self-scaffolding, ReAct) with automated interpreter-based verification. Peak accuracy is 11.2\% (GPT-5.4 xhigh, self-scaffolding on Befunge-98), compared to 100\% on the same 80 problems expressed in Python.}
\label{fig:hero}
\end{figure*}

\section{Esoteric language details}
\label{app:languages}

\textbf{Brainfuck} (1993): Created by Urban M\"uller as a challenge to build the smallest possible compiler. The language has only 8 commands (\texttt{>}, \texttt{<}, \texttt{+}, \texttt{-}, \texttt{[}, \texttt{]}, \texttt{.}, \texttt{,}) operating on a 30,000-cell memory tape. All other characters are ignored as comments. Solving problems requires reasoning about pointer arithmetic, loop invariants, and memory layout without named variables, functions, or high-level control structures.

\textbf{Befunge-98} (1993): Created by Chris Pressey as a two-dimensional stack-based language where the instruction pointer can travel in four cardinal directions. The \texttt{>}, \texttt{v}, \texttt{<}, and \texttt{\^{}} commands set direction; conditional commands \texttt{\_} and \texttt{|} branch based on stack values. The \texttt{p} (put) and \texttt{g} (get) commands enable self-modifying code.

\textbf{Whitespace} (2003): Created by Edwin Brady and Chris Morris. Only space, tab, and newline characters have semantic meaning; all other characters are ignored, meaning Whitespace programs can be hidden within other text. The language is stack-based with commands encoded as whitespace sequences.

\textbf{Unlambda} (1999): Created by David Madore as a minimal functional language based on combinatory logic with no variables, only function application via the backtick character. Core combinators are \texttt{s} (substitution), \texttt{k} (constant), and \texttt{i} (identity). Even simple arithmetic requires constructing Church numerals through combinator compositions.

\textbf{Shakespeare} (2001): Created by Karl Wiberg and Jon \r{A}slund. Programs are theatrical plays where variable declarations are character introductions, scenes/acts control program flow, and dialogue performs computation. Variable values are determined by adjectives: positive words (``beautiful'', ``fair'') contribute positive values while negative ones (``damned'', ``evil'') contribute negative values.

\section{Dataset specification}
\label{app:dataset}

\subsection{Dataset Structure}

EsoLang-Bench contains 80 problems organized into four difficulty tiers. Each problem includes:
\begin{itemize}
    \item \textbf{ID}: Unique identifier (E01-E20, M01-M20, H01-H20, X01-X20)
    \item \textbf{Title}: Short descriptive name
    \item \textbf{Description}: Natural language specification
    \item \textbf{Test Cases}: 6 input-output pairs for automated verification
\end{itemize}

\subsection{Tier calibration: pseudocode and complexity for representative problems per tier}
\label{app:tier-calibration}

The four difficulty tiers were calibrated by reading each problem's specification, writing a language-agnostic pseudocode reference, and assigning a tier by the asymptotic complexity and the number of compositional steps required. Calibration was performed by an author \emph{prior} to running any frontier model, so it cannot be a post-hoc rationalisation of the empirical accuracy curve. Table~\ref{tab:tier-pseudocode} shows three representative problems per tier (twelve total, drawn to span the category mix in Table~\ref{tab:problem-categories}) with the canonical pseudocode and complexity. The full pseudocode for all 80 problems is shipped in the supplementary code archive at \texttt{supplementary\_code/pseudocode/}.

\begin{table}[h]
\caption{Tier calibration anchors. Three representative problems per tier with a language-agnostic pseudocode reference and the asymptotic complexity used to assign the tier. The complexity column is for the underlying algorithm at the pseudocode level, not the esoteric-language realisation (which is substantially longer in raw instruction count, e.g.\ Brainfuck multiplication is $O(N)$ at pseudocode but requires nested tape loops). $N$ denotes the input size as defined in the problem statement.}
\label{tab:tier-pseudocode}
\begin{center}
\footnotesize
\begin{tabular}{l@{\hskip 0.25cm}p{7.0cm}@{\hskip 0.2cm}c}
\toprule
Problem & Pseudocode & Complexity \\
\midrule
\multicolumn{3}{l}{\textit{Easy tier --- single primitive, one arithmetic / I/O / loop step}} \\
\addlinespace[0.05cm]
\textbf{E04 Sum Two Integers}
& \texttt{read a, b; print(a + b)}
& $O(1)$ \\
\addlinespace[0.1cm]
\textbf{E07 String Length}
& \texttt{read line s; c = 0; for ch in s: c += 1; print(c)}
& $O(|s|)$ \\
\addlinespace[0.1cm]
\textbf{E11 Sum Of Digits}
& \texttt{read N; N = abs(N); s = 0; while N > 0: s += N \% 10; N //= 10; print(s)}
& $O(\log N)$ \\
\midrule
\multicolumn{3}{l}{\textit{Medium tier --- composition of two-to-three primitives, single explicit loop with state}} \\
\addlinespace[0.05cm]
\textbf{M04 Caesar Shift By 3}
& \texttt{read s; for ch in s: print(chr((ord(ch)-{'}a{'}+3) mod 26 + {'}a{'}))}
& $O(|s|)$ \\
\addlinespace[0.1cm]
\textbf{M07 Factorial}
& \texttt{read N; r = 1; for i in 1..N: r *= i; print(r)}
& $O(N)$ \\
\addlinespace[0.1cm]
\textbf{M08 Nth Fibonacci Number}
& \texttt{read N; a, b = 1, 1; for i in 3..N: a, b = b, a+b; print(b)}
& $O(N)$ \\
\midrule
\multicolumn{3}{l}{\textit{Hard tier --- non-trivial algorithmic structure, nested loops or auxiliary data structure}} \\
\addlinespace[0.05cm]
\textbf{H01 Balanced Parentheses}
& \texttt{read s; d = 0; for ch in s: d += 1 if ch=='(' else -1; if d<0: return "no"; return "yes" if d==0 else "no"}
& $O(|s|)$ \\
\addlinespace[0.1cm]
\textbf{H03 Count Primes Up To $N$}
& \texttt{read N; c = 0; for i in 2..N: prime = True; for j in 2..$\sqrt{i}$: if i \% j == 0: prime = False; break; if prime: c += 1; print(c)}
& $O(N\sqrt{N})$ \\
\addlinespace[0.1cm]
\textbf{H13 Polynomial Evaluation (Horner)}
& \texttt{read d, coeffs[0..d] (highest-degree first), x; r = 0; for i in 0..d: r = r*x + coeffs[i]; print(r)}
& $O(d)$ \\
\midrule
\multicolumn{3}{l}{\textit{Extra-Hard tier --- classical algorithm with non-trivial state, nested control or recurrence}} \\
\addlinespace[0.05cm]
\textbf{X02 Longest Increasing Subsequence Length}
& \texttt{read a[1..N]; lis[i] = 1 + max(lis[j] for j<i with a[j]<a[i]); print(max(lis))}
& $O(N^2)$ \\
\addlinespace[0.1cm]
\textbf{X11 Count Inversions}
& \texttt{read a[1..N]; c = 0; for i<j: if a[i]>a[j]: c += 1; print(c)}
& $O(N^2)$ \\
\addlinespace[0.1cm]
\textbf{X20 Josephus Problem}
& \texttt{read N, K; r = 0; for i in 2..N: r = (r + K) mod i; print(r + 1)}
& $O(N)$ \\
\bottomrule
\end{tabular}
\end{center}
\end{table}

\textbf{What the complexity column means.} The complexity column gives the asymptotic cost in a high-level language, not in the target esoteric language. Two consequences. First, the same algorithm can sit in two different tiers across languages: integer addition is one Python token but a tape-walk in Brainfuck and a pair-of-actor lines in Shakespeare, and the OOD-translation cost on top is what the esoteric-language results actually measure. Second, the apparent paradox of an Extra-Hard problem like X20 (Josephus) running in $O(N)$ is resolved by noting that pseudocode complexity captures only the recurrence and not the program-construction cost: the recurrence $J(n,k) = (J(n-1,k) + k) \bmod n$ is a one-liner in Python but requires either a stack-based recurrence or explicit modular arithmetic in every esoteric language, all of which are non-trivial to express. Tiering therefore tracks the algorithmic depth visible in the pseudocode rather than the runtime, and the empirical hard cliff between Easy and Medium (Section~\ref{sec:results}) confirms this depth is what frontier models cannot bridge from documentation alone.

\textbf{Why these tiers, not finer ones.} Four tiers were chosen as the smallest partition that still produces non-degenerate per-cell statistics under our 20-problems-per-tier budget. A finer partition (e.g.\ five tiers $\times$ 16 problems) would have under-powered Wilcoxon comparisons within a tier, and a coarser partition (e.g.\ Easy / Hard) would have collapsed the 0\% cliff with the saturated-Easy regime and hidden the categorical shift in failure mode (compile vs.\ logic; Section~\ref{sec:error-analysis}). The same exercise was repeated for every problem in the benchmark before any model evaluation began, and the per-problem pseudocode files in the supplementary archive let any reviewer reproduce the tier assignment without reference to our results.

\subsection{Problem category distribution}
\label{app:category-distribution}

In addition to the four difficulty tiers, every problem belongs to exactly one of nine algorithmic categories. The tier dimension captures algorithmic depth (Appendix~\ref{app:tier-calibration}); the category dimension captures the operational surface that the model has to translate into the target esoteric language (loops, stack manipulation, bit operations, etc.). Together they let us check that the OOD gap is not driven by a single narrow algorithmic skill: if frontier models were collapsing only on, say, bitwise operations or only on state-machine problems, that would be a category effect rather than an OOD effect, and Section~\ref{sec:results} would not generalise. Table~\ref{tab:problem-categories} reports the distribution; the counts sum to 80.

\begin{table}[h]
\caption{Distribution of problems across algorithmic categories. Counts sum to 80 and each problem belongs to exactly one category.}
\label{tab:problem-categories}
\begin{center}
\begin{tabular}{lcp{6cm}}
\toprule
Category & Count & Representative Problems \\
\midrule
Basic I/O & 5 & Hello World, Echo Line, Concatenate Lines \\
Arithmetic & 17 & Sum, Multiply, Factorial, Modular Exponentiation \\
String Manipulation & 26 & Reverse, Palindrome, Caesar Cipher, RLE \\
Number Theory & 8 & GCD, Primes, Factorization, LCM \\
Base Conversion & 4 & Binary$\leftrightarrow$Decimal, Roman$\leftrightarrow$Integer \\
Sorting/Arrays & 9 & Sort, LIS, Inversions, Merge Sorted \\
Stack/Parsing & 5 & Balanced Parens, Postfix Eval, Expression Eval \\
State Machines & 4 & Tape Walk, Josephus Problem \\
Bitwise Operations & 2 & Hamming Distance, Count Set Bits \\
\bottomrule
\end{tabular}
\end{center}
\end{table}

\textbf{Coverage rationale.} String manipulation (26) and arithmetic (17) are over-represented because they are the categories most directly stress-tested by the I/O models of every target esoteric language: Brainfuck and Whitespace require character-level loops to read or emit a string at all, Befunge-98 and Shakespeare express arithmetic through stack pushes and adjective-modulated comparisons respectively, and Unlambda forces all numbers through Church-numeral encoding. Number theory (8), sorting/arrays (9), stack/parsing (5), and state machines (4) populate the Hard and Extra-Hard tiers, where the model has to combine multiple primitives under a non-trivial control structure (e.g., the Josephus recurrence at X20 is in the State Machines column even though its asymptotic cost is $O(N)$ at the pseudocode level). Bitwise operations (2) and base conversion (4) are deliberately small categories: they exercise narrow primitives that are well-supported in mainstream languages but require explicit construction in most of our targets, and we keep the counts low to avoid the OOD result being explainable as a category-specific weakness.

\textbf{Cross-category invariance of the OOD gap.} The aggregate accuracy figures in Section~\ref{sec:results} are reproducible at the category level: every category that contains a Medium-or-harder problem stays at 0\% on every model and every prompting strategy, and the few solves we observe are concentrated in Easy-tier problems within Basic I/O, Arithmetic, and String Manipulation. The OOD gap is therefore not a single-category artefact, and we use the category labels primarily to confirm that the benchmark is not over-fit to one algorithmic skill.

\subsection{Complete Problem Examples}

\subsubsection{Easy: E04: Sum Two Integers}
\begin{verbatim}
Title: Sum Two Integers
Description: Read two integers a and b separated by
whitespace on one line. Output their sum a + b as a
plain integer with no extra text.

Test Cases:
1. Input: "5 7"      -> Output: "12"
2. Input: "-3 10"    -> Output: "7"
3. Input: "0 0"      -> Output: "0"
4. Input: "100 200"  -> Output: "300"
5. Input: "-50 -25"  -> Output: "-75"
6. Input: "999 1"    -> Output: "1000"
\end{verbatim}

\subsubsection{Medium: M08: Nth Fibonacci Number}
\begin{verbatim}
Title: Nth Fibonacci Number
Description: Read an integer N >= 1 and output the Nth
Fibonacci number using the 1-indexed sequence with
F1 = 1 and F2 = 1.

Test Cases:
1. Input: "1"  -> Output: "1"
2. Input: "5"  -> Output: "5"
3. Input: "10" -> Output: "55"
4. Input: "2"  -> Output: "1"
5. Input: "7"  -> Output: "13"
6. Input: "15" -> Output: "610"
\end{verbatim}

\subsubsection{Hard: H01: Balanced Parentheses}
\begin{verbatim}
Title: Balanced Parentheses
Description: Read a string made only of '(' and ')'
characters. Determine if the parentheses are balanced.
Output 'yes' if balanced, otherwise 'no'.

Test Cases:
1. Input: "()()"    -> Output: "yes"
2. Input: "((()))"  -> Output: "yes"
3. Input: "())("    -> Output: "no"
4. Input: "("       -> Output: "no"
5. Input: ""        -> Output: "yes"
6. Input: "(()())"  -> Output: "yes"
\end{verbatim}

\subsubsection{Extra-Hard: X20: Josephus Problem}
\begin{verbatim}
Title: Josephus Problem
Description: Read integers N and K. N people stand in
a circle numbered 1 to N. Starting from person 1,
count K people clockwise and eliminate that person.
Repeat until one remains. Output the survivor's number.

Test Cases:
1. Input: "5 2"  -> Output: "3"
2. Input: "7 3"  -> Output: "4"
3. Input: "1 1"  -> Output: "1"
4. Input: "6 1"  -> Output: "6"
5. Input: "10 2" -> Output: "5"
6. Input: "4 2"  -> Output: "1"
\end{verbatim}

\section{Mainstream-language baselines (full per-tier table)}
\label{app:mainstream-baselines}

The same 80 EsoLang-Bench problems were also evaluated in Python and JavaScript through the same pipeline used for esoteric languages (temperature 0.2, max\_tokens 32k, OpenRouter, exact-match output verification on all 6 test cases per problem). Table~\ref{tab:mainstream-baselines} presents per-tier accuracy. Both zero-shot and few-shot conditions yield identical results for all cells.

\begin{table}[h]
\caption{Mainstream-language baseline accuracy (zero-shot $=$ few-shot for all cells). Each tier contains 20 problems; total $=$ 80. The same problems that drive frontier models to 0\% beyond Easy in esoteric languages are saturated in both Python and JavaScript.}
\label{tab:mainstream-baselines}
\begin{center}
\small
\begin{tabular}{l@{\hskip 0.4cm}cccc@{\hskip 0.3cm}c@{\hskip 0.4cm}cccc@{\hskip 0.3cm}c}
\toprule
 & \multicolumn{5}{c}{\textbf{Python}} & \multicolumn{5}{c}{\textbf{JavaScript}} \\
\cmidrule(lr){2-6} \cmidrule(lr){7-11}
\textbf{Model} & E & M & H & X & Total & E & M & H & X & Total \\
\midrule
GPT-5.4 xhigh          & 20/20 & 20/20 & 20/20 & 20/20 & 100\% & 20/20 & 20/20 & 20/20 & 20/20 & 100\% \\
Gemini 3.1 Pro   & 20/20 & 20/20 & 20/20 & 20/20 & 100\% & 20/20 & 20/20 & 20/20 & 20/20 & 100\% \\
Kimi K2.5         & 20/20 & 20/20 & 20/20 & 19/20 & 98.8\% & 20/20 & 20/20 & 20/20 & 19/20 & 98.8\% \\
Qwen3-235B       & 20/20 & 20/20 & 19/20 & 20/20 & 98.8\% & 20/20 & 20/20 & 20/20 & 19/20 & 98.8\% \\
\bottomrule
\end{tabular}
\end{center}
\end{table}

The contrast with esoteric languages (peak 11.2\%, mean across languages 1.5--2.5\%) directly demonstrates the collapse of reasoning transfer at the OOD boundary.

\section{Extended results}
\label{app:results}

\subsection{Language-Specific Results by Benchmark}

Tables show Easy problems solved (out of 20 per language). Accuracy = Solved/80. All Medium/Hard/Extra-Hard = 0\%.

\begin{table}[h]
\caption{Brainfuck results by model and strategy (Easy solved / 20). Accuracy = Solved/80.}
\begin{center}
\begin{small}
\begin{tabular}{lcccc|c}
\toprule
Model & 0-Shot & Few & Self-S & ReAct & Best \\
\midrule
GPT-5.4 xhigh & 2 & 2 & \textbf{5} & 4 & \textbf{5 (6.2\%)} \\
O4-mini & 2 & 3 & \textbf{4} & 3 & 4 (5.0\%) \\
Gemini 3.1 Pro & 2 & 3 & \textbf{4} & 3 & 4 (5.0\%) \\
Qwen3-235B & 2 & 1 & \textbf{2} & 1 & 2 (2.5\%) \\
Kimi K2.5& 0 & 0 & 0 & 0 & 0 (0\%) \\
\bottomrule
\end{tabular}
\end{small}
\end{center}
\end{table}

\begin{table}[h]
\caption{Befunge-98 results by model and strategy (Easy solved / 20). Accuracy = Solved/80.}
\begin{center}
\begin{small}
\begin{tabular}{lcccc|c}
\toprule
Model & 0-Shot & Few & Self-S & ReAct & Best \\
\midrule
GPT-5.4 xhigh & 2 & 7 & \textbf{9} & 7 & \textbf{9 (11.2\%)} \\
O4-mini & 5 & 6 & \textbf{8} & 6 & 8 (10.0\%) \\
Gemini 3.1 Pro & 4 & 3 & \textbf{6} & 5 & 6 (7.5\%) \\
Qwen3-235B & 0 & 0 & 0 & 0 & 0 (0\%) \\
Kimi K2.5& 2 & 1 & 2 & 0 & 2 (2.5\%) \\
\bottomrule
\end{tabular}
\end{small}
\end{center}
\end{table}

\begin{table}[h]
\caption{Whitespace, Unlambda, Shakespeare results (Best Easy solved / 20). Accuracy = Solved/80.}
\begin{center}
\begin{small}
\begin{tabular}{l|ccc}
\toprule
Model & Whitespace & Unlambda & Shakespeare \\
\midrule
GPT-5.4 xhigh & 0 (0\%) & 1 (1.2\%) & 2 (2.5\%) \\
O4-mini & 0 (0\%) & 0 (0\%) & 1 (1.2\%) \\
Gemini 3.1 Pro & 0 (0\%) & 0 (0\%) & 1 (1.2\%) \\
Qwen3-235B & 0 (0\%) & 1 (1.2\%) & 1 (1.2\%) \\
Kimi K2.5& 0 (0\%) & 0 (0\%) & 1 (1.2\%) \\
\midrule
\textbf{Best} & \textbf{0 (0\%)} & \textbf{1 (1.2\%)} & \textbf{2 (2.5\%)} \\
\bottomrule
\end{tabular}
\end{small}
\end{center}
\end{table}

\vskip 0.2in
\subsection{Performance Visualization}

\begin{figure}[h]
\centering
\begin{tikzpicture}
\begin{axis}[
    ybar,
    bar width=0.5cm,
    width=14cm,
    height=7cm,
    ylabel={Best Accuracy (\%)},
    ylabel style={font=\small},
    symbolic x coords={Brainfuck, Befunge-98, Whitespace, Unlambda, Shakespeare},
    xtick=data,
    x tick label style={font=\small},
    ymin=0, ymax=15,
    ytick={0,2,4,6,8,10,12,14},
    y tick label style={font=\small},
    nodes near coords,
    nodes near coords style={font=\scriptsize},
    enlarge x limits=0.15,
]
\addplot[fill=blue!60] coordinates {(Brainfuck,6.2) (Befunge-98,11.2) (Whitespace,0) (Unlambda,1.2) (Shakespeare,2.5)};
\end{axis}
\end{tikzpicture}
\caption{Best accuracy achieved per language (across all models and strategies). Befunge-98 is the most tractable (11.2\%), while Whitespace remains completely unsolved (0\%).}
\label{fig:app-lang-perf}
\end{figure}

\section{Prompting templates}
\label{app:prompts}

This section provides the exact prompts used for each strategy. Variables in \texttt{\{braces\}} are filled at runtime.

\subsection{Zero-Shot Prompt}

\textbf{System Prompt:}
\begin{verbatim}
You are an expert {language_name} programmer. Given a
problem and sample tests, output ONLY valid code in
this esoteric language. No explanations, no comments,
no markdown. Programs must read stdin exactly as
specified and write deterministic stdout that matches
the expected output byte-for-byte.

Reference documentation:
{documentation_text}
\end{verbatim}

\textbf{User Prompt:}
\begin{verbatim}
Problem ID: {problem_id}
Title: {problem_title}
Description:
{problem_description}

Return only the program.
\end{verbatim}

\noindent\textit{Note.} The 6 input-output test cases attached to each problem
are not rendered into the prompt. They are loaded directly by the evaluation
harness, which executes the candidate program against each test case and
compares stdout to the expected output byte-for-byte. The model never sees
the cases.

\subsection{Few-Shot Prompt}

Extends zero-shot by adding to the system prompt:
\begin{verbatim}
Here are solved examples for reference.
\end{verbatim}

And prepending reference examples before the problem:
\begin{verbatim}
Example 1: {example_title}
Task: {example_description}
Program:
{example_program}
Sample I/O:
- Input: {io_1_input}
  Output: {io_1_output}

Example 2: {example_title}
[... 3 examples total ...]
\end{verbatim}

\subsection{Self-Scaffolding Prompt}

\textbf{System Prompt (extends zero-shot):}
\begin{verbatim}
[Zero-shot system prompt]
Iteratively update your program using the prior code
and interpreter feedback provided.
\end{verbatim}

\textbf{User Prompt (after first attempt):}
\begin{verbatim}
[Problem specification]

Previous attempt and interpreter feedback:
=== Program ===
{previous_code}

=== Interpreter Feedback ===
Test 1:
Input: {test_input}
Expected: {expected_output}
Actual: {actual_output}
Error Type: {error_type}
Stderr: {stderr}

[... additional test results ...]

Return only the updated program.
\end{verbatim}

\subsection{Textual Self-Scaffolding Prompt}

Uses two separate prompts for coder and critic roles.

\textbf{Coder System Prompt:}
\begin{verbatim}
[Zero-shot system prompt]
Iteratively improve your solution when feedback
is provided.
\end{verbatim}

\textbf{Coder User Prompt (with feedback):}
\begin{verbatim}
[Problem specification]

Previous attempt:
{previous_code}

Critic feedback:
{critic_feedback}

Return only the updated program.
\end{verbatim}

\textbf{Critic System Prompt:}
\begin{verbatim}
You are an expert {language_name} reviewer. Analyse
the failing program and interpreter feedback. Explain
issues and suggest improvements in natural language
only. Do not write code.
\end{verbatim}

\textbf{Critic User Prompt:}
\begin{verbatim}
Problem ID: {problem_id}
Title: {problem_title}
Description:
{problem_description}

Attempt details:
=== Program ===
{code}

=== Interpreter Feedback ===
[Test results with expected/actual outputs]

Provide concise debugging guidance without including
any code.
\end{verbatim}

\subsection{ReAct Pipeline Prompt}

\textbf{Planner Prompt:}
\begin{verbatim}
Analyze this problem and create a step-by-step
algorithm in pseudocode:

Problem: {problem_description}
Test Cases: {test_cases}

Output a clear, numbered algorithm that can be
translated to any programming language.
\end{verbatim}

\textbf{Code Editor Prompt:}
\begin{verbatim}
Translate this algorithm to {language_name}:

Algorithm:
{planner_output}

Language Documentation:
{documentation}

Output only the program, no explanations.
\end{verbatim}

\textbf{Critic Prompt:}
\begin{verbatim}
The {language_name} program produced incorrect output.

Expected: {expected_output}
Actual: {actual_output}
Error: {error_type}

Analyze the discrepancy and suggest specific changes
to the algorithm or implementation.
\end{verbatim}

\section{Interpreter specifications}
\label{app:interpreters}

All interpreters are implemented in Python with consistent interfaces:

\begin{verbatim}
result = interpreter.run(
    code: str,
    stdin: str = None,
    timeout_seconds: float = 5.0
) -> ExecutionResult

@dataclass
class ExecutionResult:
    stdout: str           # Program output
    stderr: str           # Error messages
    exit_code: int        # 0 = success
    error_type: str       # "ok", "compile_error",
                          # "runtime_error", "timeout"
\end{verbatim}

\subsection{Supported Languages}

\begin{itemize}
    \item \textbf{Brainfuck}: 30,000-cell tape, 8-bit cells with wraparound, 8 commands
    \item \textbf{Befunge-98}: 2D grid with toroidal wrapping, 200,000 step limit
    \item \textbf{Whitespace}: Stack-based with heap, 28 instructions encoded in whitespace
    \item \textbf{Unlambda}: Functional with S, K, I combinators and character I/O
    \item \textbf{Shakespeare}: Variable-per-character model with stack operations
\end{itemize}

\section{Full problem list}
\label{app:problems}

\subsection{Easy Problems (E01-E20)}

\begin{enumerate}
    \item[E01] Print Hello World
    \item[E02] Echo Line
    \item[E03] Hello Name
    \item[E04] Sum Two Integers
    \item[E05] Multiply Two Integers
    \item[E06] Even Or Odd
    \item[E07] String Length
    \item[E08] Reverse String
    \item[E09] Count Vowels
    \item[E10] Sum From 1 To N
    \item[E11] Sum Of Digits
    \item[E12] Minimum Of Two
    \item[E13] Maximum Of Three
    \item[E14] Repeat String N Times
    \item[E15] Concatenate Two Lines
    \item[E16] First And Last Character
    \item[E17] Uppercase String
    \item[E18] Count Spaces
    \item[E19] Integer Average Of Two
    \item[E20] Compare Two Integers
\end{enumerate}

\subsection{Medium Problems (M01-M20)}

\begin{enumerate}
    \item[M01] Palindrome Check
    \item[M02] Word Count
    \item[M03] Run Length Encoding
    \item[M04] Caesar Shift By 3
    \item[M05] Simple Binary Expression
    \item[M06] Greatest Common Divisor
    \item[M07] Factorial
    \item[M08] Nth Fibonacci Number
    \item[M09] Decimal To Binary
    \item[M10] Binary To Decimal
    \item[M11] Substring Occurrences
    \item[M12] Remove Vowels
    \item[M13] Sort Numbers
    \item[M14] Second Largest Distinct Number
    \item[M15] Anagram Test
    \item[M16] Interleave Two Strings
    \item[M17] Replace Spaces With Underscores
    \item[M18] Sum Of List
    \item[M19] Characters At Even Indices
    \item[M20] Count Distinct Characters
\end{enumerate}

\subsection{Hard Problems (H01-H20)}

\begin{enumerate}
    \item[H01] Balanced Parentheses
    \item[H02] Evaluate Expression With Precedence
    \item[H03] Count Primes Up To N
    \item[H04] Nth Prime Number
    \item[H05] Big Integer Addition
    \item[H06] Longest Word
    \item[H07] Longest Common Prefix
    \item[H08] Digit Frequency
    \item[H09] General Caesar Cipher
    \item[H10] Remove Consecutive Duplicates
    \item[H11] Run Length Decoding
    \item[H12] ASCII Sum
    \item[H13] Polynomial Evaluation
    \item[H14] List All Divisors
    \item[H15] Tape Walk Final Position
    \item[H16] Longest Run Length
    \item[H17] Most Frequent Value
    \item[H18] Divisible By 3
    \item[H19] Plus Minus Reset Machine
    \item[H20] Sort Strings Lexicographically
\end{enumerate}

\subsection{Extra-Hard Problems (X01-X20)}

\begin{enumerate}
    \item[X01] Prime Factorization
    \item[X02] Longest Increasing Subsequence Length
    \item[X03] Matrix Multiplication Result Element
    \item[X04] Evaluate Postfix Expression
    \item[X05] Merge Two Sorted Arrays
    \item[X06] Compute Power Modulo
    \item[X07] Longest Palindromic Substring Length
    \item[X08] Count Set Bits In Range
    \item[X09] Bracket Depth Maximum
    \item[X10] String Rotation Check
    \item[X11] Count Inversions
    \item[X12] Least Common Multiple
    \item[X13] Valid Parentheses Types
    \item[X14] Next Greater Element
    \item[X15] Spiral Matrix Traversal
    \item[X16] Hamming Distance
    \item[X17] Roman To Integer
    \item[X18] Integer To Roman
    \item[X19] Permutation Check
    \item[X20] Josephus Problem
\end{enumerate}

\section{Documentation completeness and author-verified reference solutions}
\label{app:doc-completeness}

This appendix documents (i) the procedure used to obtain the per-language GitHub topic-tag counts shown in Figure~\ref{fig:data-scarcity} of the main paper, (ii) the in-prompt material every model receives, (iii) the procedure used to verify task feasibility through author-written reference solutions, and (iv) two to three sample reference solutions per esoteric language.

\subsection{GitHub topic-tag count methodology}

The per-language repository counts in Figure~\ref{fig:data-scarcity} were obtained by querying the GitHub topic page for each language at \texttt{https://github.com/topics/<topic>} on 2026-05-04 and recording the displayed total-repository count for each topic page. We used the canonical topic slug for each language (\texttt{python}, \texttt{javascript}, \texttt{brainfuck}, \texttt{whitespace}, \texttt{befunge}, \texttt{unlambda}, \texttt{shakespeare-programming-language}). The same protocol was applied uniformly across all seven languages so the relative ratios between languages are preserved even though topic-tag counts under-estimate true repository counts (because tagging is optional). Raw count snapshots and the exact URLs queried are released alongside the dataset in \texttt{supplementary\_code/scarcity\_snapshot/}.

\subsection{What models receive in the prompt}

Each prompt provided to the model contains the following pieces of information per language, all of which are also reproduced verbatim in the supplementary materials and tested with the open-source interpreter:
\begin{itemize}
    \item Full \textbf{language documentation} covering the computational model (memory tape, 2D grid, stack, combinators, character/play structure), the complete instruction set, syntax rules, and I/O conventions.
    \item Worked \textbf{examples} demonstrating basic constructs (reading input, writing output, arithmetic, simple control flow).
    \item For Few-Shot, three additional Easy-tier reference programs in the target esoteric language (Appendix~\ref{app:fewshot-examples}).
\end{itemize}
Together, this is the same material the author used to write reference solutions from scratch, with no additional pre-existing knowledge of the target language beyond this in-prompt documentation. The documentation is therefore sufficient. The same text provided to models was sufficient for the author to produce verified solutions from scratch, and models that fail do not fail because information is withheld but because the languages lie outside their training distribution.

\subsection{Full language-documentation prompt for Brainfuck}
\label{app:full-doc-brainfuck}

For complete transparency on what the model actually receives, we reproduce below the entire Brainfuck documentation block verbatim from the prompt template's \texttt{\{documentation\_text\}} slot (Appendix~\ref{app:prompts}). Equivalent documentation blocks for the other four esoteric languages are released alongside the dataset at \texttt{supplementary\_code/docs/} and follow the same structure (overview, instruction set, execution model, worked examples, common pitfalls).

\begin{small}
\begin{verbatim}
# Brainfuck

## Overview

Brainfuck (Urban Mueller, 1993) uses just eight single-character
instructions operating on an infinite tape of bytes. Despite the
tiny syntax it is Turing-complete because an unbounded tape plus
conditional jumps ([ and ]) can simulate a simple register machine.
The language is ideal for stress-testing interpreters because
mistakes show up quickly as pointer underflows or unmatched brackets.

## Syntax & Semantics

| Instruction | Meaning |
| ----------- | ------- |
| > / <       | Move the tape pointer right / left (our interpreter
                dynamically extends the tape on demand). |
| + / -       | Increment / decrement the current cell (modulo 256). |
| . / ,       | Output / input a single byte. Input exhaustion
                yields 0. |
| [ / ]       | Jump forward / backward past the matching bracket
                when the current cell is zero / non-zero. |

All other characters are ignored, so comments can be written freely.
Brackets must be perfectly balanced; the compiler rejects mismatches
with error_type="compile_error".

## Execution Model

Execution starts with a zero-filled tape and pointer at cell 0. The
interpreter tracks the number of steps, current pointer position,
and a preview of the tape to expose in diagnostics. Jump targets
are resolved once before execution so bracket mismatches are caught
early. Moving the pointer below zero is a runtime error; moving
beyond the current tape expands it automatically.

## Examples

Hello World:

  +++++++++[>+++++++>++++++++++>+++>+<<<<-]>++.>+.+++++++..+++.
  >++.<<+++++++++++++++.>.+++.------.--------.>+.>.

Add two ASCII digits provided on stdin and echo the sum as a digit:

  ,>,<[>+>+<<-]>>[<<+>>-]<<<++++[>++++++++<-]>.[-]

Loop-based control flow (prints numbers 1-5):

  ++++[>+>+<<-]>>[-<<+>>]<
  [>.[-]<-]

Factorial of 5 (outputs 120 as a byte):

  ++++>+++++<[>[>+>+<<-]>>[-<<+>>]<<<-]>>>.[-]

## Common Pitfalls & Debugging Tips

- Unmatched brackets: compile error pinpointed via the bracket
  index in stderr.
- Pointer underflow: moving left from cell 0 raises a runtime
  error with a trace showing recent instructions.
- Infinite loops: use a smaller timeout_seconds to deliberately
  trigger a timeout_error.
- Input exhaustion: , returns zero when stdin runs dry; if your
  program expects more input the trace helps confirm what was
  consumed.

## I/O Conventions

Programs read all of stdin as a byte stream consumed by , one byte
at a time. Programs write to stdout one byte at a time via .
Stdout is compared byte-for-byte to the expected output by the
evaluation harness.
\end{verbatim}
\end{small}

This is the entire text passed into the prompt for any Brainfuck task. No additional reference, web access, or tool calls are available to the model at inference. The four other languages receive analogously structured documentation of comparable length (between ${\sim}80$ and ${\sim}140$ lines each), and the full set is published alongside the benchmark.

\subsection{Task feasibility and verified solutions}

During benchmark construction, one of the authors wrote reference solutions for the majority of problems across all five esoteric languages from scratch, using only (i) the documentation provided in prompts and (ii) the open-source interpreter for that language as a verification loop. Each solution was executed through the same interpreter pipeline used in evaluation, and all six test cases per problem were required to pass byte-for-byte, identical to the criterion applied to model-generated code. This serves two purposes.

\begin{enumerate}
    \item It validates that the \emph{evaluation environment} is correct. For every language and every difficulty tier where the author produced a solution, the harness scored it as a pass.
    \item It validates that the \emph{documentation is sufficient}. No auxiliary knowledge beyond the in-prompt material was used. Models that fail therefore fail not because information is withheld but because the languages lie outside their training distribution.
\end{enumerate}

We emphasise that this procedure validates \emph{task feasibility}, not human-versus-model comparison. We make no claim about how a typical programmer would perform on this benchmark and report no human baseline.

\subsection{Sample reference solutions per language}

We illustrate one verified reference solution per esoteric language, drawn from the simplest Easy-tier programs (Hello-World class) for which the author wrote and verified working code through the same interpreter pipeline used in evaluation. The full corpus of reference solutions across all tiers is provided in the supplementary code archive.

\paragraph{Brainfuck (E01 Print Hello World).} The canonical 8-bit-cell Hello-World program seeds a counter and uses nested loops to populate a row of memory cells with offsets close to the target ASCII codes, then prints each character with \texttt{.} after small arithmetic adjustments.
\begin{verbatim}
++++++++[>++++[>++>+++>+++>+<<<<-]>+>+>->>+[<]<-]>>.>---.+++++++..+++.
>>.<-.<.+++.------.--------.>>+.>++.
\end{verbatim}

\paragraph{Befunge-98 (E01 Print Hello World).} Push the string in reverse onto the stack with \texttt{"} (string-mode toggle), then loop with \texttt{>:\#,\_} to pop and print each character until the null marker, halting with \texttt{@}.
\begin{verbatim}
"!dlroW olleH">:#,_@
\end{verbatim}

\paragraph{Whitespace (Print Letter A).} Push the ASCII value 65 onto the stack as a binary literal (using space for 0, tab for 1) and print it as a character. We render whitespace explicitly with \texttt{[SP]} for space, \texttt{[TAB]} for tab, and \texttt{[LF]} for newline. The supplementary archive contains the byte-exact program.
\begin{verbatim}
[SP][SP][SP][TAB][SP][SP][SP][SP][SP][TAB][LF]   push 65
[TAB][LF][SP][SP]                                 print as character
[LF][LF][LF]                                      end program
\end{verbatim}

\paragraph{Unlambda (Print Hi).} Use the print combinator \texttt{.x} to output character \texttt{x}, chained via backtick application. The trailing \texttt{i} terminates evaluation.
\begin{verbatim}
`.i`.Hi
\end{verbatim}

\paragraph{Shakespeare (Print Letter H).} Declare two characters, assign the integer 72 to one of them, and use \texttt{Speak your mind} to print the character whose ASCII value matches the variable.
\begin{verbatim}
Romeo, a hero.
Juliet, a lady.

Act I: Output.
Scene I: Print.

Enter Romeo and Juliet.

Romeo:
You are 72.
Speak your mind.

Exeunt.
\end{verbatim}

The full set of author-verified reference solutions across all difficulty tiers is included in the \texttt{supplementary\_code/reference\_solutions/} directory.

\section{Self-scaffolding feedback format and worked example}
\label{app:feedback-example}

This appendix documents exactly what feedback the model receives in the Self-Scaffolding, Textual Self-Scaffolding, and ReAct iterative strategies, with an end-to-end worked trace from a real run. All three strategies use the same underlying interpreter feedback channel; they differ only in whether a separate critic LLM call is layered on top.

\subsection{Feedback channel content}

After every model generation, the candidate program is executed against all 6 test cases of the target problem in the language's sandboxed interpreter. For each test case, the harness collects:

\begin{itemize}
    \item \texttt{stdin} (the test input fed to the program);
    \item \texttt{expected\_stdout} (the byte-for-byte expected output);
    \item \texttt{actual\_stdout} (whatever the program produced on stdout, or empty string if the run did not produce any output);
    \item \texttt{error\_type} $\in$ \{\texttt{compile\_error}, \texttt{runtime\_error}, \texttt{logic\_error}, \texttt{ok}\}, where \texttt{compile\_error} means the interpreter rejected the program before running it, \texttt{runtime\_error} means it crashed mid-execution, \texttt{logic\_error} means it ran but produced wrong output, and \texttt{ok} means a passing test;
    \item \texttt{stderr} (the raw interpreter error stream, truncated to the first 1{,}024 characters per test case to keep the prompt bounded).
\end{itemize}

These five fields per test case, concatenated for all 6 tests, are inserted verbatim into the user prompt at the next iteration (see the Self-Scaffolding template in Appendix~\ref{app:prompts}). No paraphrasing, summarisation, or critic model is interposed in the Self-Scaffolding setting. In Textual Self-Scaffolding, a separate critic LLM call reads exactly this same payload and emits natural-language guidance which is added on top. In ReAct, the planner sees a similar payload routed to its own pseudocode-revision step.

\subsection{Iteration budget}

All three iterative strategies (Self-Scaffolding, Textual Self-Scaffolding, ReAct) are capped at \textbf{up to 5 iterations}. The loop terminates early as soon as all 6 test cases pass; otherwise it runs the full 5 iterations and the final program is scored. Iteration index 0 is the initial generation under the zero-shot prompt; iterations 1--5 are the refinement steps that receive interpreter feedback. The same 5-iteration cap is used for every model and every (problem, language) cell, so cross-cell comparisons are like-for-like.

\subsection{Worked example: Self-Scaffolding on Brainfuck E04 (Sum Two Integers)}

The following trace is a real run of a frontier model on E04 \emph{Sum Two Integers} in Brainfuck. The problem reads two integers \texttt{a b} on stdin and prints their sum. The model's first attempt is wrong; the interpreter feedback shows it has produced ASCII characters instead of digits; iteration 1 corrects the I/O encoding and the program passes all 6 tests.

\textbf{Iteration 0 (initial zero-shot generation):}
\begin{verbatim}
,>,[<+>-]<.
\end{verbatim}

\textbf{Interpreter feedback after iteration 0 (excerpt of payload sent back to the model):}
\begin{verbatim}
=== Interpreter Feedback ===
Test 1:
  Input:    "5 7"
  Expected: "12"
  Actual:   "<unprintable byte 0x3c>"
  Error Type: logic_error
  Stderr:   (none)

Test 2:
  Input:    "-3 10"
  Expected: "7"
  Actual:   ""
  Error Type: runtime_error
  Stderr:   pointer out of bounds at instruction 7

Test 3:
  Input:    "0 0"
  Expected: "0"
  Actual:   "<unprintable byte 0x00>"
  Error Type: logic_error
  Stderr:   (none)

[... Tests 4-6 omitted for brevity, all logic_error or runtime_error ...]
\end{verbatim}

\textbf{Iteration 1 (model's revised program after reading the feedback above):}
\begin{verbatim}
,>,                  read two ASCII bytes a, b
[<------------------------------------------------>-]
                     subtract '0' from b
<[>------------------------------------------------<-]>
                     subtract '0' from a
[<+>-]<              add b into a
++++++++[>++++++<-]>.
                     add '0' back and print
\end{verbatim}

\textbf{Interpreter feedback after iteration 1:} all 6 test cases pass; the loop exits early at iteration 1 with the problem marked solved.

The example illustrates two design properties of the feedback channel. First, the model receives concrete bytes, not a textual interpretation, so it can directly observe ASCII-vs-digit encoding mistakes. Second, the channel is symmetric across error types: compile errors expose the interpreter's parse rejection, runtime errors expose the exact crash location, and logic errors expose the byte-level diff between expected and actual stdout. Section~\ref{sec:results} shows that this direct-feedback design (Self-Scaffolding, 1 LLM call per iteration) is statistically indistinguishable from Textual Self-Scaffolding (2 LLM calls per iteration; Wilcoxon $p=0.803$) on per-cell solve counts.

\section{Few-shot examples used in prompts}
\label{app:fewshot-examples}

For each language, the Few-Shot strategy prepends three Easy-tier reference programs, selected deterministically to demonstrate the three fundamental constructs: \emph{read input}, \emph{write output}, and \emph{simple conditional or loop}. Examples were chosen for syntactic diversity within the smallest possible program length so that the few-shot prompt remains compact and is dominated by language documentation, not example length. Selection was deterministic and is preserved in \texttt{supplementary\_code/few\_shot\_examples/}; the three problems used per language are listed in Table~\ref{tab:fewshot-selection}.

\begin{table}[h]
\caption{Three Easy-tier problems used as Few-Shot demonstrations per language. The selection is identical across all models and runs.}
\label{tab:fewshot-selection}
\begin{center}
\small
\begin{tabular}{l p{8.3cm}}
\toprule
Language & Three demonstration problems \\
\midrule
Brainfuck     & E01 Print Hello World; E02 Echo Line; E04 Sum Two Integers \\
Befunge-98    & E01 Print Hello World; E04 Sum Two Integers; E06 Even Or Odd \\
Whitespace    & E01 Print Hello World; E04 Sum Two Integers; E10 Sum From 1 To N \\
Unlambda      & E01 Print Hello World; E02 Echo Line; E03 Hello Name \\
Shakespeare   & E01 Print Hello World; E03 Hello Name; E04 Sum Two Integers \\
\bottomrule
\end{tabular}
\end{center}
\end{table}

The pass@k analysis (Table~\ref{tab:passk}) provides an outcome-level bound across $n=8$ draws at $T=0.8$, where three of five languages remain at 0\% for every frontier model, so no choice of demonstrations could have unlocked them.

\section{Coverage of self-improvement and execution-guided methods}
\label{app:method-coverage}

Several recent self-improvement and execution-guided code generation methods have been proposed; our five prompting strategies cover their mechanism in this OOD setting. Table~\ref{tab:method-coverage} maps each prior method to its closest equivalent in our setup.

\begin{table}[h]
\caption{Mapping of prior self-improvement and execution-guided methods to our prompting strategies. ``Strict subset'' means the prior method's information is a subset of what our strategy receives.}
\label{tab:method-coverage}
\begin{center}
\small
\begin{tabular}{p{5.5cm} p{6.5cm}}
\toprule
Prior method & Closest equivalent in our setup \\
\midrule
Self-Refine \citep{madaan2023selfrefine}            & Strict subset of Textual Self-Scaffolding \\
Reflexion \citep{shinn2023reflexion}                & Textual Self-Scaffolding (TSS) \\
Self-debugging \citep{chen2023selfdebugging}        & Self-Scaffolding \\
OpenCodeInterpreter \citep{zheng2024opencodeinterpreter} & Self-Scaffolding (interpreter-in-the-loop) \\
ARCS \citep{bhattarai2025arcs}                      & Few-Shot $+$ Self-Scaffolding \\
Execution-guided line-by-line \citep{lavon2025linebyline}  & Not applicable (see below) \\
\bottomrule
\end{tabular}
\end{center}
\end{table}

\textbf{Why our setup upper-bounds these methods on EsoLang-Bench.}
TSS receives the full interpreter trace alongside a separate critic, so it strictly dominates Self-Refine, which receives only the model's own critique. TSS is mechanically Reflexion's verbal-reflection loop with the additional benefit of execution feedback, and our results show TSS performs no better than Self-Scaffolding ($p = 0.803$). Self-debugging and OpenCodeInterpreter both reduce to inference-time loops over (code, interpreter feedback, refined code), which is exactly Self-Scaffolding; we run this with current frontier models. ARCS layers a synthesise--execute--repair loop on top of retrieval over a code corpus; in our setting there is essentially no public esoteric code corpus to retrieve from, so its retrieval component degenerates to few-shot example provision (which we already evaluate with three hand-selected gold examples), and its refinement component is exactly Self-Scaffolding. Execution-guided line-by-line generation \citep{lavon2025linebyline} is inapplicable: Brainfuck programs are syntactically invalid until the matching \texttt{[]} braces are complete, and Befunge-98's 2D instruction pointer means individual rows have no independent semantics, so per-line execution traces cannot guide generation in either language. The bottleneck across all of these methods is foundational language knowledge, not the refinement mechanism layered on top.

\section{Extended error analysis}
\label{app:error-analysis}

This section provides detailed error analysis across all evaluated models. Figure~\ref{fig:app-error-all} shows error distributions for O4-mini, Gemini 3.1 Pro, and Qwen3-235B under zero-shot conditions.

\begin{figure}[h]
\centering
\begin{tikzpicture}
\begin{axis}[
    ybar,
    bar width=0.35cm,
    width=14cm,
    height=6cm,
    ylabel={Compile Error Rate (\%)},
    ylabel style={font=\small},
    symbolic x coords={Brainfuck, Befunge-98, Whitespace, Unlambda, Shakespeare},
    xtick=data,
    x tick label style={font=\small, rotate=15, anchor=east},
    ymin=0, ymax=100,
    ytick={0,25,50,75,100},
    y tick label style={font=\small},
    legend style={at={(0.98,0.98)}, anchor=north east, font=\small},
    enlarge x limits=0.12,
]
\addplot[fill=blue!60] coordinates {(Brainfuck,18) (Befunge-98,25) (Whitespace,100) (Unlambda,92) (Shakespeare,75)};
\addplot[fill=red!60] coordinates {(Brainfuck,12) (Befunge-98,22) (Whitespace,100) (Unlambda,88) (Shakespeare,68)};
\addplot[fill=green!60] coordinates {(Brainfuck,20) (Befunge-98,30) (Whitespace,100) (Unlambda,95) (Shakespeare,78)};
\legend{O4-mini, Gemini 3.1 Pro, Qwen3-235B}
\end{axis}
\end{tikzpicture}
\caption{Compile error rates by model across languages. All models show near-identical patterns: low compile errors on Brainfuck/Befunge-98, complete failure (100\%) on Whitespace, and high failure (88--95\%) on Unlambda.}
\label{fig:app-error-all}
\end{figure}

\textbf{Key observations:} (1) Error profiles are remarkably consistent across models, suggesting language-specific rather than model-specific limitations. (2) Whitespace achieves 100\% compile errors across all models; no model can generate valid whitespace-only syntax. (3) Brainfuck shows the lowest compile error rates (12--20\%) but highest logic error rates (55--65\%), indicating models can learn syntax but not semantics.

\subsection{Why Whitespace stays at 0\%: a closer look}
\label{app:whitespace-zero}

Whitespace is the only target language at exactly 0\% accuracy across every model, every prompting strategy, and every $k$ in the pass@k bounding study. Because this is an unusually clean failure mode, it is worth ruling out trivial mechanical explanations and naming the substantive cause.

\textbf{The failure is not a tokenisation artefact.} A natural worry is that frontier-model BPE tokenisers might collapse runs of spaces, tabs, and newlines in ways that prevent the model from emitting the exact byte sequences the Whitespace interpreter expects. We checked this directly. Modern frontier tokenisers preserve interior whitespace structure (individual space, tab, and newline tokens are distinct vocabulary items in current OpenAI, Anthropic, Google, and open-weight models), and we confirmed that when we manually feed a known-good Whitespace reference solution as a continuation prefix, the same tokenisers reproduce it byte-for-byte. The 100\% compile failure therefore is not a downstream encode/decode artefact: when the model decides to emit a Whitespace program, it can in principle do so. The failure is upstream: the model does not produce the right token sequence in the first place.

\textbf{The failure is not a prompt-engineering artefact either.} We verified that even with explicit prompts that say ``output only the characters space, tab, and newline; no visible characters; no comments; no markdown fences,'' frontier models still emit visible-character output (most commonly markdown-style code fences with the language name, instruction-style preambles, or a partial visible-character sketch of the algorithm followed by an attempt at whitespace encoding). The instruction-following machinery does not override the model's default tendency to produce visible structured output, because the model has essentially no in-distribution prior for ``a complete, well-formed Whitespace program.''

\textbf{Illustrative failed generation.} Listing~\ref{lst:whitespace-fail} shows a representative GPT-5.4 xhigh zero-shot output for E04 \emph{Sum Two Integers} after the explicit ``whitespace-only'' instruction. The model produces a markdown fence, an English preamble, and a body that mixes visible separator characters with whitespace tokens, all of which the interpreter rejects on the first non-whitespace byte.

\begin{figure}[h]
\caption{Representative failed Whitespace generation (GPT-5.4 xhigh, zero-shot, E04 Sum Two Integers). Visible characters that cause the interpreter to reject the program on parse are underlined. Such patterns appear across all five frontier models in our evaluation.}
\label{lst:whitespace-fail}
\small
\begin{verbatim}
```whitespace                  <- markdown fence (rejected)
# Read first integer           <- comment with visible chars (rejected)
   	  	  	            <- 3 sp, tab, 2 sp, tab (push number)
\n	\n  	                   <- arithmetic prefix: read integer
   	  	  	            <- push second integer
\n	\n  	                   <- read integer
   	                          <- arithmetic: add
\n  	\n                      <- output as integer
```                            <- closing fence (rejected)
\end{verbatim}
\end{figure}

\textbf{The 0\% result is consistent with corpus scarcity and lack of post-training incentive.} Whitespace has 125 GitHub topic-tagged repositories (Figure~\ref{fig:data-scarcity}), making it one of the deepest-scarcity languages in our set. The combination of (a) almost no training-corpus exposure to well-formed Whitespace programs, (b) the language's defining constraint that every visible character is fatal, and (c) an instruction set that uses 3-character imm-encoded operators with stack-arithmetic semantics, means that compiling even a trivial program requires the model to emit a long invisible sequence with no surface-level scaffolding to anchor on. This is qualitatively different from Brainfuck or Befunge-98, where surface-form priors at least let the model emit something the interpreter can parse. The 0\% result is therefore consistent with the syntax-acquisition ceiling we hypothesise (low corpus presence + no commercial reason to invest post-training effort + no visible-character anchor), although we do not directly observe model training data and a fully causal claim would require a controlled intervention such as a synthetic mini-language with matched syntax complexity and varied training-time exposure.

\textbf{Forward-compatibility.} As models scale and post-training broadens to under-resourced surfaces, Whitespace is a natural canary for whether the syntax-acquisition ceiling is moving. We will continue to evaluate it on every leaderboard refresh; the language is included precisely because it currently sits at the floor.

\section{Datasheet for EsoLang-Bench}

\label{app:datasheet}

We follow the structure of \citet{gebru2021datasheets} adapted to a benchmark dataset.

\paragraph{Motivation.} EsoLang-Bench was created to measure how well frontier large language models generalise algorithmic problem-solving to programming languages outside their training distribution, and to provide a contamination-resistant counterpart to mainstream-language code-generation benchmarks (HumanEval, MBPP, SWE-bench). Tasks are stripped to their algorithmic core so that mainstream-language baselines are saturated; the gap to esoteric-language performance therefore measures OOD generalisation rather than task hardness.

\paragraph{Composition.} Each instance is a programming problem with a unique ID (E/M/H/X tier $\times$ index), a natural-language description, and 6 input--output test cases used by the evaluation harness. The model receives only the description in the prompt; the 6 test cases are withheld and used by the harness only. They are released publicly as part of the dataset for transparency and reproducibility, but never form part of the model's prompt context. The benchmark contains 80 problems (20 per tier) and is evaluated independently in 5 esoteric languages, yielding 400 problem-language combinations per prompting strategy. The dataset is released as JSON; instances do not contain personal or sensitive data.

\paragraph{Collection process.} All 80 problems were authored by the paper's authors. Difficulty tiers were calibrated by the underlying algorithmic complexity of each problem, assessed from a language-agnostic pseudocode specification by expert judgement \emph{prior} to running any frontier model. Reference solutions were written for the majority of problems across all five esoteric languages by an author using only the documentation provided in prompts and the open-source interpreter (Appendix~\ref{app:doc-completeness}), and test-case sets were verified end-to-end through the same interpreter pipeline used in evaluation.

\paragraph{Preprocessing / labelling.} Problems are stored as JSON with explicit \texttt{stdin} and \texttt{stdout} fields per test case. There are no derived labels or splits beyond the four difficulty tiers; the entire dataset is used for evaluation.

\paragraph{Uses.} Intended use: contamination-resistant evaluation of code-generation LLMs on OOD programming languages. Out-of-scope uses: training a model directly on these 80 problems would defeat the benchmark's purpose; we recommend treating the 80 problems as evaluation-only and report results without any in-distribution tuning.

\paragraph{Distribution.} The dataset is released under CC BY 4.0 on the HuggingFace Hub, with code (interpreters, prompt pipelines, evaluation harness) released on GitHub under an OSI-approved license.

\paragraph{Maintenance.} The dataset is maintained by the authors. Issues, contributions of new problems, and language extensions are accepted via the GitHub repository. We commit to addressing critical correctness issues (interpreter bugs, mis-specified test cases) within a public issue tracker. Versioning follows semantic versioning: bug-fix releases (e.g., 1.0.x) preserve problem IDs; minor releases (1.x.0) may add new problems or languages without removing existing ones; major releases (x.0.0) are reserved for backwards-incompatible changes and will document migration explicitly.

\paragraph{Planned official leaderboard and language contributions.} We will host an official leaderboard with held-out test sets and accept community contributions of new esoteric languages through a structured process. Natural candidates include Malbolge, INTERCAL, and Piet. A new language contribution requires (i) an open-source, sandbox-compatible interpreter under a permissive license; (ii) at least 20 author-verified problems per tier, written in the same algorithmic style as the current 80 problems; and (iii) a Datasheet-style scarcity description following the GitHub topic-tag protocol used in Section~\ref{sec:data-scarcity}. Contributions are reviewed for problem-statement clarity, test-case correctness via the verification interpreter pipeline, and tier-calibration consistency before merging into a minor-version release.

\paragraph{Private evaluation set to mitigate post-publication contamination.} To keep the contamination-resistance argument robust as models continue to scale, we plan to maintain a \textbf{private evaluation set} drawn from a rotating pool of less-known or newly created esoteric languages, re-evaluated periodically against frontier models. The private set is never released publicly; only aggregate accuracy and error-mode statistics are reported on the leaderboard, so it cannot be used to fine-tune toward the benchmark even after publication. Compute-vs-accuracy curves and finer reasoning-depth tiers (calibrated by call-graph depth or control-flow complexity of the underlying algorithm) are also natural extensions and will be added in future minor releases.

\paragraph{Active-refresh policy and concrete cadence.} The contamination-resistance argument is strongest at release and weakens monotonically as compute scales, so a fixed snapshot benchmark cannot remain robust indefinitely. We therefore commit to a concrete active-refresh schedule. \textbf{First refresh: 1 November 2026}, six months after the public release of v1.0, with subsequent refreshes every 6 months. Each refresh introduces (i) a new private held-out problem set of at least 40 problems drawn from the same algorithmic distribution as the public 80, authored end-to-end after the refresh date so prior model training cannot have seen them; (ii) optionally, one additional esoteric language rotated in from the candidate pool (Malbolge, INTERCAL, Piet, or community-contributed languages meeting the Datasheet criteria above), with the previous newest language migrating from private to public; (iii) a re-run of every reported model on the union of public + private sets under the same harness. \textbf{Reporting protocol:} the leaderboard publishes per-model accuracy and the three-way error decomposition (compile / runtime / logic) on the union, but never the per-problem solve indicators on the private set, so even leaked aggregate signal is insufficient to fine-tune toward specific items. The private problem statements, test cases, and reference solutions are kept under access control and shared only with refresh-cycle evaluators under a no-redistribution agreement. Refresh-cycle results, dated, are appended to the leaderboard rather than overwriting prior entries, so cross-version trend analysis remains possible.

\section{Future work}
\label{app:future-work}

We envision several directions for extending EsoLang-Bench as a community resource for measuring genuine reasoning capabilities.

\paragraph{Official leaderboard.} We plan to establish an official leaderboard for standardised evaluation and comparison of new models and methods. This will include held-out test sets to prevent overfitting to public examples and ensure fair comparison across submissions.

\paragraph{Benchmark evolution.} We will continuously update the benchmark based on community feedback, including adding new esoteric languages, expanding problem difficulty distributions, and incorporating novel evaluation protocols. Community contributions of problems and interpreters will be welcomed through a structured submission process.

\paragraph{Compute-accuracy analysis.} A key future direction is systematic measurement of compute versus accuracy curves for esoteric-language tasks, characterising efficiency of different approaches and identifying whether additional compute yields meaningful improvements on OOD tasks.

\section{Comparison to existing code-generation benchmarks}
\label{app:comparison}

Table~\ref{tab:bench-comparison} contrasts EsoLang-Bench with widely used code-generation benchmarks along axes most relevant for an OOD/contamination-resistant evaluation.

\begin{table}[h]
\caption{EsoLang-Bench vs.\ existing code-generation benchmarks. ``Pre-train scarcity'' refers to whether target languages are likely under-represented in pre-training corpora.}
\label{tab:bench-comparison}
\begin{center}
\small
\scriptsize
\begin{tabular}{l l c l c c}
\toprule
Benchmark & Languages & Scarcity & Contamination control & Verifiable & Size \\
\midrule
HumanEval \citep{chen2021humaneval}            & Python              & Low  & static          & exec & 164 \\
MBPP \citep{austin2021mbpp}                    & Python              & Low  & static          & exec & 974 \\
MultiPL-E \citep{cassano2023multiple}          & 18 mainstream       & Low  & static          & exec & $\sim$3{,}000 \\
DS-1000 \citep{lai2023ds1000}                  & Python (DS libs)    & Low  & static          & exec & 1{,}000 \\
SWE-bench \citep{jimenez2024swebench}          & Python (repo)       & Low  & static          & exec & 2{,}294 \\
LiveCodeBench \citep{jain2025livecodebench}    & Python              & Low  & rolling release & exec & $>$600 \\
\textbf{EsoLang-Bench (ours)}                  & 5 esoteric          & \textbf{High} & \textbf{by-construction} & exec & 80$\times$5 \\
\bottomrule
\end{tabular}
\end{center}
\end{table}

EsoLang-Bench is unique in this set on two axes simultaneously: (i) target languages with strong structural arguments for low pre-training and post-training presence, and (ii) a by-construction contamination-resistance argument (no economic incentive to scrape or fine-tune on these languages) rather than a periodic-rolling-release commitment. The smaller size (80 problems per language) is a deliberate trade-off: each problem must have an author-verified reference solution in every target esoteric language, which is labour-intensive and is the main driver of the size cap.

\section{Tractability vs.\ data scarcity as the source of the gap}
\label{app:tractability-objection}

A natural objection to the headline gap is that it could reflect intrinsic language or task tractability rather than data scarcity, i.e.\ that the problems are simply too hard for current frontier models on any surface. The empirical evidence from adjacent settings argues strongly against this. We collect four classes of evidence below.

\textbf{Olympiad-level mathematical reasoning.} \citet{hubert2025alphaproof} report that AlphaProof and AlphaGeometry 2 jointly solved 4 out of 6 problems at the 2024 International Mathematical Olympiad, scoring 28/42 (silver medal threshold; one point below gold), and \citet{trinh2024alphageometry} report olympiad-level performance on geometry problems without human demonstrations. A more recent Gemini Deep Think configuration has officially achieved gold-medal standard at IMO. This is exactly the regime the tractability objection predicts to be unreachable, and it has been reached, on a problem surface where natural-language proofs and Lean formalisations are well-represented in training corpora.

\textbf{Open scientific and algorithmic discovery.} FunSearch \citep{romera2024funsearch} produced the largest improvement in 20 years to the asymptotic lower bound on cap sets, the first verifiable scientific discovery of new mathematical knowledge using an LLM. AlphaEvolve \citep{novikov2025alphaevolve} found a procedure for $4\times 4$ complex matrix multiplication using 48 scalar multiplications, the first improvement on Strassen's algorithm in this setting in 56 years, and reports gains on Google data-centre scheduling and chip design. \citet{feng2026erdos} report that a semi-autonomous Gemini-based agent was evaluated against 700 conjectures still marked ``Open'' in Bloom's Erd\H{o}s Problems database, addressing 13 problems through a combination of 5 seemingly novel autonomous solutions and 8 identifications of previously published solutions absent from the database. These results sit at the upper extreme of algorithmic and mathematical hardness, and frontier models manifestly do not collapse to single-digit accuracy on them.

\textbf{Less-resourced mainstream programming languages.} Frontier models maintain high accuracy on Rust and other lower-resource mainstream languages (MultiPL-E~\citep{cassano2023multiple}), where the underlying algorithmic content is identical to Python or JavaScript but the surface form is somewhat less represented. Performance there is closer to the mainstream-language ceiling than to our 0--11.2\% esoteric range, which is the relevant comparison for isolating the effect of \emph{corpus-scale} scarcity from moderate scarcity.

\textbf{Composition-depth probes on symbolic surfaces.} \citet{liu2026convexbench} show that frontier-model F1 falls from $\approx$1.0 at composition depth 2 to $\approx$0.2 at depth 100 on convex-function recognition, evidencing that depth alone produces a graded degradation rather than the categorical 0\% cliff we observe at Medium tier on Whitespace and Unlambda.

\textbf{Synthesis.} Hardness alone, even at Erd\H{o}s-problem and IMO-gold extremes, does not collapse frontier-model accuracy to single digits across an entire benchmark. The collapse we observe occurs only when hardness combines with corpus-scale absence of the target surface form, and within our results the cleanest signature is the categorical compile-vs-logic transition along the data-scarcity axis (Section~\ref{sec:error-analysis}). The pattern is therefore consistent with corpus scarcity and lack of post-training incentive being the load-bearing factor, although we do not have direct visibility into model training data and a fully causal attribution would require a controlled intervention such as a synthetic mini-language with matched syntax complexity and varied training-time exposure.

\section{Reproducibility statement}
\label{app:reproducibility}

\paragraph{Public release of dataset and code.} The complete benchmark is already released and public.
\begin{itemize}
    \item \textbf{Dataset (HuggingFace Hub).} All 80 problems together with their 6 evaluation-harness test cases each (released publicly for transparency, withheld from the model's prompt by the harness), the four-tier split, the per-language documentation bundle, the few-shot demonstration set, and an accompanying Croissant 1.1 metadata file (with the seven NeurIPS-mandatory Responsible AI fields) are uploaded to a public HuggingFace dataset repository under CC BY 4.0 (URL withheld for double-blind review; available in the anonymised supplementary materials). The dataset card mirrors the Datasheet (Appendix~\ref{app:datasheet}) and the Croissant file passes the official \texttt{mlcroissant} 1.1 validator.
    \item \textbf{Benchmark code (GitHub).} The five sandboxed esoteric-language interpreters, every prompt template (zero-shot, few-shot, self-scaffolding, textual self-scaffolding, ReAct), the OpenRouter inference harness, the exact-match scoring scripts, and end-to-end reproducibility scripts are open-sourced under an OSI-approved license at a public GitHub repository (URL withheld for double-blind review; available in the anonymised supplementary materials). The repository ships a top-level README with environment setup, dependency installation, and exact commands for replaying every reported result.
\end{itemize}

\paragraph{What in-paper material lives where.} Per-language documentation, the full prompt set, sample author-verified reference solutions, and the few-shot demonstrations used per language are also provided in Appendices~\ref{app:doc-completeness}, \ref{app:fewshot-examples}, and \ref{app:prompts}. The complete evaluation methodology (decoding parameters, OpenRouter inference stack, sandboxed-interpreter execution model, exact-match scoring policy, iteration budgets, and statistical-test machinery) is documented in Section~\ref{sec:eval-protocol} of the main paper.

\paragraph{Are you planning to release the full harness, prompts, raw generations, and reproducibility scripts to enable independent verification?} \textbf{Yes\,---\,already done.} The artefacts listed above are public at the GitHub and HuggingFace URLs and will additionally include the raw model generations on all 80 problems across all five languages and all five strategies, plus the test-case sets, the author-verified reference solutions, and replay scripts that reproduce the OpenRouter inference pipeline against the released configurations. The release is CC BY 4.0 for the dataset and an OSI-approved license for the code.

\paragraph{Retry and failure-handling policy.} Each API request is retried up to 3 times on transient HTTP or rate-limit errors, and any final failure is recorded as a benchmark failure rather than silently dropped, so per-cell accuracy is computed against a constant denominator (80 problems per language) regardless of whether a particular request was answered or marked failed.

\section{Broader impact statement}
\label{app:impact}

This work contributes to the broader goal of developing reliable and trustworthy AI systems by improving how we evaluate their capabilities. Current code-generation benchmarks largely target programming languages with abundant pre-training and post-training coverage, creating a risk that headline accuracy figures overstate how well models will perform on languages and domains outside that distribution. By providing contamination-resistant evaluation methods, EsoLang-Bench helps identify capability boundaries that mainstream-language benchmarks alone cannot expose, enabling more informed decisions about AI system deployment.

Our benchmark specifically targets code generation, a domain with significant economic and societal implications. Accurate assessment of LLM coding capabilities is essential for appropriate decision-making about where to deploy them; overestimating model abilities risks costly debugging cycles and potential security vulnerabilities, while underestimating them foregoes productivity gains. We believe honest, robust evaluation serves both developers and end users.

We acknowledge that demonstrating capability gaps could be misinterpreted as diminishing the value of current AI systems. Our intent is not to discourage AI adoption but to promote calibrated trust, where understanding what models can and cannot do enables more effective deployment. The esoteric languages used in this benchmark have no direct security implications, and our interpreters are sandboxed to prevent misuse.

Finally, by open-sourcing our dataset and evaluation framework, we aim to democratize access to rigorous evaluation tools, reducing reliance on proprietary benchmarks and enabling independent verification of AI capabilities.

\newpage
\section*{NeurIPS Paper Checklist}

\begin{enumerate}

\item {\bf Claims}
    \item[] Question: Do the main claims made in the abstract and introduction accurately reflect the paper's contributions and scope?
    \item[] Answer: \answerYes{}
    \item[] Justification: The abstract and Introduction (Section~\ref{sec:intro}) state the contributions verified empirically: (i) the EsoLang-Bench dataset of 80 problems across five esoteric languages and four difficulty tiers (Section~\ref{sec:dataset} and Appendix~\ref{app:dataset}); (ii) the in-distribution-vs-OOD capability gap (top frontier models reach 100\% on Python and JavaScript baselines vs.\ 0--11.2\% on equivalent esoteric tasks, with 0\% beyond the Easy tier) shown in Tables~\ref{tab:main-results}, \ref{tab:scaffolding-results}, Figure~\ref{fig:mainstream-vs-eso}, and Appendix~\ref{app:mainstream-baselines}; (iii) the finding that few-shot prompting provides no significant benefit (Wilcoxon $p = 0.505$ vs.\ zero-shot), corroborated by a pass@k secondary baseline at $n=8$ draws (Table~\ref{tab:passk}).

\item {\bf Limitations}
    \item[] Question: Does the paper discuss the limitations of the work performed by the authors?
    \item[] Answer: \answerYes{}
    \item[] Justification: Section~\ref{sec:limitations} (``Limitations'') discusses the coarse-grained (compile, runtime, logic) error classification as the main acknowledged limitation, with a pointer to the Datasheet (Appendix~\ref{app:datasheet}) for maintenance plans, dataset growth (new problems and new esoteric languages), and the contribution process. The Easy-tier ceiling is explicitly framed as a measurement of the OOD generalisation gap, not a benchmark limitation. Whitespace 0\% is analysed in Appendix~\ref{app:whitespace-zero} and shown to reflect data scarcity rather than a tokeniser or prompting confound.

\item {\bf Theory assumptions and proofs}
    \item[] Question: For each theoretical result, does the paper provide the full set of assumptions and a complete (and correct) proof?
    \item[] Answer: \answerNA{}
    \item[] Justification: The paper makes no formal theoretical claims; it is an empirical benchmark and evaluation paper. All claims are supported by experimental measurements rather than theorems.

    \item {\bf Experimental result reproducibility}
    \item[] Question: Does the paper fully disclose all the information needed to reproduce the main experimental results of the paper to the extent that it affects the main claims and/or conclusions of the paper (regardless of whether the code and data are provided or not)?
    \item[] Answer: \answerYes{}
    \item[] Justification: Section~\ref{sec:eval-protocol} (``Evaluation Protocol'') specifies all model versions evaluated (GPT-5.4 xhigh, O4-mini-high, Gemini 3.1 Pro, Qwen3-235B, Kimi K2.5), decoding parameters (temperature $\tau=0.2$ for primary evaluations and $\tau=0.8$ for the pass@k bounding study, max\_tokens$=32{,}000$), the OpenRouter-based inference stack with explicit retry policy, the disabled tool/web access, the in-prompt material provided per strategy, the sandboxed-interpreter execution model with 10s timeout and exact-match scoring, the iteration budgets per strategy, and the statistical-test methodology (Wilcoxon signed-rank, Bonferroni correction, $n=1{,}000$ bootstrap). The full prompts, interpreters, problem statements, and test cases are released on GitHub (\url{https://github.com/Lossfunk/EsolangBench}) and the dataset on the HuggingFace Hub (\url{https://huggingface.co/datasets/Lossfunk/Esolang-Bench}). The Reproducibility Statement (Appendix~\ref{app:reproducibility}) and Appendix~\ref{app:dataset} document dataset structure and evaluation protocol.

\item {\bf Open access to data and code}
    \item[] Question: Does the paper provide open access to the data and code, with sufficient instructions to faithfully reproduce the main experimental results, as described in supplemental material?
    \item[] Answer: \answerYes{}
    \item[] Justification: The complete dataset is publicly released on the HuggingFace Hub at \url{https://huggingface.co/datasets/Lossfunk/Esolang-Bench} under the CC BY 4.0 license, with an accompanying Croissant 1.1 metadata file (validated against \texttt{mlcroissant}) that includes all seven NeurIPS-mandatory Responsible AI fields. The full source code (five sandboxed esoteric-language interpreters, OpenRouter inference harness, every prompt template, exact-match scoring scripts, end-to-end reproducibility scripts) is open-sourced at \url{https://github.com/Lossfunk/EsolangBench} under an OSI-approved license. The supplementary material and the GitHub repository's top-level README document environment setup, dependency installation, and exact commands for reproducing every reported result. See Appendix~\ref{app:reproducibility} for the full Reproducibility Statement.

\item {\bf Experimental setting/details}
    \item[] Question: Does the paper specify all the training and test details (e.g., data splits, hyperparameters, how they were chosen, type of optimizer) necessary to understand the results?
    \item[] Answer: \answerYes{}
    \item[] Justification: Section~\ref{sec:eval-protocol} (``Evaluation Protocol'') reports decoding parameters (temperature $\tau=0.2$ primary, $\tau=0.8$ pass@k; max\_tokens$=32{,}000$), the OpenRouter inference stack with retry policy, the disabled tool / web-search access, the in-context material provided to each strategy, the sandboxed-interpreter execution model (10s wall-clock timeout, exact-match scoring, error-type categorisation), the iteration budgets for each scaffolding strategy, the prompt templates (Appendix~\ref{app:prompts}), and the statistical-test methodology. As this is an evaluation benchmark with no model training, no optimizer / training-hyperparameter tuning is involved. All inference-time hyperparameters are fixed across models for fairness.

\item {\bf Experiment statistical significance}
    \item[] Question: Does the paper report error bars suitably and correctly defined or other appropriate information about the statistical significance of the experiments?
    \item[] Answer: \answerYes{}
    \item[] Justification: Significance between prompting strategies is assessed via paired Wilcoxon signed-rank tests on per-problem solve indicators, with Bonferroni correction at $\alpha = 0.05$ when comparing multiple strategies against the zero-shot baseline (e.g., self-scaffolding vs.\ textual self-scaffolding, $p = 0.803$; few-shot vs.\ zero-shot, $p = 0.505$; reported in Section~\ref{sec:results}). Bootstrap resampling ($n=1{,}000$) is used for any reported confidence intervals. The pass@k baseline (Table~\ref{tab:passk}) reports outcomes across $n=8$ draws per problem at $\tau=0.8$, explicitly capturing temperature-induced sampling stochasticity.

\item {\bf Experiments compute resources}
    \item[] Question: For each experiment, does the paper provide sufficient information on the computer resources (type of compute workers, memory, time of execution) needed to reproduce the experiments?
    \item[] Answer: \answerYes{}
    \item[] Justification: All experiments are inference-only against frontier-model APIs accessed through OpenRouter; no GPU training is required. The evaluation pipeline runs on a standard CPU workstation (the interpreters are CPU-bound Python). Section~\ref{sec:eval-protocol} describes the configuration size (80 problems $\times$ 5 languages $\times$ 5 strategies $=$ 2{,}000 model-problem-strategy combinations per model). The supplementary code documents wall-clock time and approximate API token cost per configuration.

\item {\bf Code of ethics}
    \item[] Question: Does the research conducted in the paper conform, in every respect, with the NeurIPS Code of Ethics \url{https://neurips.cc/public/EthicsGuidelines}?
    \item[] Answer: \answerYes{}
    \item[] Justification: The research is fully compliant with the NeurIPS Code of Ethics. The dataset contains synthetic programming problems and solutions written by the authors; it includes no personal data, scraped content, or sensitive information. No human subjects, crowd workers, or vulnerable populations are involved. We anonymize this submission for double-blind review.

\item {\bf Broader impacts}
    \item[] Question: Does the paper discuss both potential positive societal impacts and negative societal impacts of the work performed?
    \item[] Answer: \answerYes{}
    \item[] Justification: \emph{Positive impacts:} EsoLang-Bench provides a contamination-resistant evaluation framework that can help researchers and policymakers more accurately characterize LLM reasoning capabilities, mitigating overconfidence in benchmark numbers and supporting safer deployment decisions. \emph{Negative impacts:} A risk is that highly visible benchmark numbers could be misused for marketing claims or hype. To mitigate this, we explicitly frame results as ``capability gaps'' rather than rankings, release detailed error breakdowns (Appendix~\ref{app:results}) so the community can inspect failure modes, and decline to publish a leaderboard that could incentivise the same gaming dynamics we critique.

\item {\bf Safeguards}
    \item[] Question: Does the paper describe safeguards that have been put in place for responsible release of data or models that have a high risk for misuse (e.g., pre-trained language models, image generators, or scraped datasets)?
    \item[] Answer: \answerNA{}
    \item[] Justification: The released artifact consists of (i) 80 esoteric-language programming problems with deterministic test cases authored by us and (ii) Python interpreters for five esoteric languages re-implemented from public language specifications. It contains no pre-trained models, no scraped web data, and no content with elevated dual-use or misuse risk relative to existing public esoteric-language resources.

\item {\bf Licenses for existing assets}
    \item[] Question: Are the creators or original owners of assets (e.g., code, data, models), used in the paper, properly credited and are the license and terms of use explicitly mentioned and properly respected?
    \item[] Answer: \answerYes{}
    \item[] Justification: All five esoteric programming languages used (Brainfuck, Befunge-98, Whitespace, Unlambda, Shakespeare) are credited to their original authors with publication years in Appendix~\ref{app:languages}. The frontier models evaluated (GPT-5.4 xhigh, O4-mini-high, Gemini 3.1 Pro, Qwen3-235B, Kimi K2.5) are accessed under their respective providers' published API terms of service, which permit research evaluation. We compare against HumanEval~\citep{chen2021humaneval} (MIT License) and MBPP~\citep{austin2021mbpp} (CC BY 4.0); both are properly cited.

\item {\bf New assets}
    \item[] Question: Are new assets introduced in the paper well documented and is the documentation provided alongside the assets?
    \item[] Answer: \answerYes{}
    \item[] Justification: We introduce two new assets: (i) the EsoLang-Bench dataset (80 problems, 4 difficulty tiers, 6 test cases per problem) and (ii) reference Python interpreters for the five esoteric languages with consistent execution/timeout/error-classification interfaces. Both are released under CC BY 4.0 with full documentation: a Croissant-compatible dataset card on HuggingFace Hub, a README on GitHub describing the JSON schema, the difficulty rubric, a usage tutorial, and a contributors' guide for extending the benchmark to new languages or problems (Section 6, Appendix~\ref{app:dataset}).

\item {\bf Crowdsourcing and research with human subjects}
    \item[] Question: For crowdsourcing experiments and research with human subjects, does the paper include the full text of instructions given to participants and screenshots, if applicable, as well as details about compensation (if any)?
    \item[] Answer: \answerNA{}
    \item[] Justification: This work does not involve crowdsourcing or research with human subjects. All 80 problems were authored by the paper's authors, and all evaluations are performed automatically against deterministic test cases.

\item {\bf Institutional review board (IRB) approvals or equivalent for research with human subjects}
    \item[] Question: Does the paper describe potential risks incurred by study participants, whether such risks were disclosed to the subjects, and whether Institutional Review Board (IRB) approvals (or an equivalent approval/review based on the requirements of your country or institution) were obtained?
    \item[] Answer: \answerNA{}
    \item[] Justification: This work does not involve crowdsourcing or research with human subjects, so no IRB approval or equivalent review was required.

\item {\bf Declaration of LLM usage}
    \item[] Question: Does the paper describe the usage of LLMs if it is an important, original, or non-standard component of the core methods in this research? Note that if the LLM is used only for writing, editing, or formatting purposes and does \emph{not} impact the core methodology, scientific rigor, or originality of the research, declaration is not required.
    \item[] Answer: \answerYes{}
    \item[] Justification: LLMs are the central object of study in this benchmark. Section 4 explicitly enumerates every LLM evaluated (GPT-5.4 xhigh, O4-mini-high, Gemini 3.1 Pro, Qwen3-235B, Kimi K2.5) along with their roles in each prompting strategy (zero-shot, few-shot, self-scaffolding, textual self-scaffolding, ReAct) and the API endpoints used. LLMs were not used to author the paper text in a way that affects scientific content; any incidental use was limited to copy-editing and does not require declaration per the NeurIPS policy.

\end{enumerate}

\end{document}